\documentclass[machines,article,accept,pdftex,moreauthors]{Definitions/mdpi} 
\firstpage{1} 
\pubvolume{1}
\issuenum{1}
\articlenumber{0}
\pubyear{2025}
\copyrightyear{2025}
\datereceived{ } 
\daterevised{ } 
\dateaccepted{ } 
\datepublished{ } 
\hreflink{https://doi.org/} 

\usepackage{array}

\newcolumntype{+}{!{\vrule width 2pt}}

\newlength\savedwidth

\newcommand\thickhline{\noalign{\global\savedwidth\arrayrulewidth\global\arrayrulewidth 2pt}%
\hline
\noalign{\global\arrayrulewidth\savedwidth}}

\usepackage{multirow}

\Title{Monocular Vision-Based Swarm Robot Localization Using Equilateral Triangular Formations}

\TitleCitation{Monocular Vision-Based Swarm Robot Localization Using Equilateral Triangular Formations}

\Author{Taewon Kang $^{1}$, Ji-Wook Kwon $^{2}$, Il Bae $^{1}$ and Jin Hyo Kim $^{3,}$*}

\AuthorNames{Taewon Kang, Ji-Wook Kwon, Il Bae and Jin Hyo Kim}

\isAPAStyle{%
       \AuthorCitation{Kang, T., Kwon, J.-W., Bae, I., \& Kim, J. H.}
         }{%
        \isChicagoStyle{%
        \AuthorCitation{Kang, Taewon, Kwon Ji-Wook, Bae Il and Kim Jin Hyo.}
        }{
        \AuthorCitation{Kang, T.; Kwon, J.-W.; Bae, I.; Kim, J. H.}
        }
}

\address{%
$^{1}$ \quad School of Integrated Technology, Yonsei University, Incheon 21983, Republic of Korea; taewon.kang@yonsei.ac.kr (T.K.); kakao@yonsei.ac.kr (I.B.)\\
$^{2}$ \quad Korea Institute of Robotics and Technology Convergence, Pohang 37666, Republic of Korea; jwkwon@kiro.re.kr (J.-W.K.)\\
$^{3}$ \quad DOGU, Inc., San Jose, CA 95134, USA; ceo@dogu.xyz (J.H.K.)}

\corres{Correspondence: ceo@dogu.xyz}

\abstract{Localization of mobile robots is crucial for deploying robots in real-world applications such as search and rescue missions.
This work aims to develop an accurate localization system applicable to swarm robots equipped only with low-cost monocular vision sensors and visual markers.
The system is designed to operate in fully open spaces, without landmarks or support from positioning infrastructures.
To achieve this, we propose a localization method based on equilateral triangular formations.
By leveraging the geometric properties of equilateral triangles, the accurate two-dimensional position of each participating robot is estimated using one-dimensional lateral distance information between robots, which can be reliably and accurately obtained with a low-cost monocular vision sensor.
Experimental and simulation results demonstrate that, as travel time increases, the positioning error of the proposed method becomes significantly smaller than that of a conventional dead-reckoning system, another low-cost localization approach applicable to open environments.
}


\begin{document}

\section{Introduction}

Localization is a fundamental research area for enabling the autonomy of mobile robots \citep{Pahlavan20153058}.
Even with a high-performance controller, a robot cannot reach its intended destination if its position estimate is inaccurate \citep{Fink2013290}.
Accurate and reliable localization systems are therefore essential for autonomous robotic operation.

Various positioning systems that use infrastructure and onboard sensors have been employed to determine the location of mobile robots.
Infrastructure-based systems---such as the Global Positioning System (GPS) \citep{Enge11, Tamazin201677, Park21:Single, Jia21:Ground, Lee22:Urban, Kim23:Single, Lee23:Seamless, Kim23:Machine} and indoor positioning systems \citep{Liu20071067, Lee20:Neural, Lee22:Evaluation, Chen201524595, Guerra2016, Kang20:Practical, Kim23:Low, Moon24:HELPS}---estimate position based on time-of-flight (TOF) or received signal strength (RSS) measurements of radio signals, along with the known locations of infrastructure components such as satellites or beacons.
These systems typically offer high accuracy but become ineffective when the infrastructure is unavailable---for example, during disasters, radio frequency interference, or emergency construction.
In particular, GPS is susceptible to ionospheric anomalies \citep{Lee22:Optimal, Lee17:Monitoring, Sun20:Performance, Sun21:Markov, Ahmed17:Statistical} and radio frequency interference \citep{Grant2009173, Park18:Dual, Chen12:Design, Lee22:SFOL, Kim19:Mitigation, Chen10:Real, JubaerAlam2019, Kim22:First}.

Positioning systems based on onboard sensors, such as dead-reckoning systems that utilize a wheel encoder and an inertial measurement unit \citep{Jirawimut2003209, Cho20112907, Byun2019, Kim2023:Balancing}, have also been employed by mobile robots.
In these systems, robots acquire their position information independently of external positioning infrastructures.
However, such systems suffer from low positioning accuracy due to accumulated and diverging errors over time.
Map-based positioning algorithms have also been investigated to improve localization accuracy using onboard sensors \citep{Filliat2003243, GaminiDissanayake2001229, Tang2017}.
Given a predefined obstacle map of the environment, a mobile robot estimates its position by matching detected obstacles to the map.
Onboard obstacle detection sensors, such as laser range finders and TOF cameras, are typically used for this purpose.
However, the positioning accuracy of map-based systems significantly degrades in large open spaces without obstacles (e.g., lobbies or long straight hallways).
Moreover, these methods cannot be applied in unknown environments lacking an existing map.

To overcome the dependency on an existing map, simultaneous localization and mapping (SLAM) algorithms have been developed, in which map building and localization processes are performed simultaneously \citep{Thrun200252, Durrant-Whyte200699, Bresson20151827, Munguia2016, Xu2025:Virtual}.
With SLAM, localization is possible even in unknown environments without positioning infrastructures or a preexisting map; consequently, SLAM has become one of the most prominent research topics in addressing the localization problem.
However, to acquire accurate obstacle information necessary for map building, SLAM typically relies on high-performance distance-measuring sensors, such as high-cost light detection and ranging (lidar) devices \citep{Forster2017249, Mu2020157628, Daoud2018, Aznar2014, Chen2023:Overview, Shao2023:Advancing}.
Moreover, the environment must contain sufficiently distinctive features or patterns to construct a high-quality map.
To compensate for accumulated localization errors in SLAM, a mobile robot’s trajectory should include loops that introduce additional travel distance; scan matching through loop closure can then improve position accuracy \citep{Williams20091188, Newman2005635, Wang2021}.
In environments such as straight passages without loop closures, however, the robot may produce a curved map, resulting in significant inaccuracy---even in a long, straight corridor.

Cooperative positioning systems, which are applicable to swarm robots, have been proposed to reduce positioning errors in uncharted environments \citep{Kurazume19941250, Kim17:Simulation}.
However, these systems typically assume that the robots are equipped with high-cost ranging sensors capable of providing accurate inter-robot distance measurements.
In other words, if the robots lack such high-precision distance-measuring sensors, the cooperative positioning methods in \citep{Kurazume19941250, Kim17:Simulation} cannot achieve high localization accuracy.

In this work, our objective is to achieve high localization accuracy for swarm robots operating in wide open spaces without landmarks.
The robots are equipped only with low-cost monocular vision sensors and visual markers.
It is assumed that there is no infrastructure support, no existing map, and no distance-measuring sensors available.
Furthermore, sophisticated vision processing is avoided to minimize computational cost.

To achieve this goal, we propose a cooperative localization system based on equilateral triangular formations.
A minimum of four robots cooperate in the system, although it can be easily extended to accommodate more robots.
In the proposed approach, three robots act as beacons and form an equilateral triangle.
The fourth robot, referred to as the moving robot, approaches the target vertex of the next equilateral triangle---closer to the destination---with the assistance of the three beacon robots.
The key challenge in implementing this system is guiding the moving robot to the target vertex with high accuracy using only low-cost monocular vision sensors and the visual markers on the beacon robots.
We address this challenge by relying solely on the lateral distance information between the beacon robots, which can be reliably obtained with high accuracy even using low-cost monocular vision sensors.
The geometric properties of equilateral triangles are also leveraged in the implementation.

Assuming that the initial positions of the robots are known (an assumption commonly made in previous studies on dead-reckoning and Kalman-filter-based SLAM, e.g., \citep{Anousaki199942, Guivant2001242, Mourikis20061273, Poulose2019}), the triangular formation advances repeatedly toward the destination, and the robots estimate their positions in succession.
A simple path planning algorithm is also developed to avoid collisions.
The detailed implementation of the system, along with its performance evaluated through experiments and simulations, is discussed in this paper.

The contributions of this study are summarized as follows:
\begin{itemize}
\item Unlike infrastructure-based localization systems, which are inoperable in environments lacking positioning infrastructures, the proposed triangular formation-based approach operates independently of any preconstructed infrastructure.

\item Unlike map-based localization systems that require existing obstacle maps, high-cost distance-measuring sensors, and significant computational resources, the proposed algorithm does not rely on obstacle maps or map matching and incurs low computational cost.
It remains operational even in open spaces without any landmarks.

\item Unlike dead-reckoning systems, where positioning errors accumulate continuously over time, in the proposed method, errors accumulate only when a new triangular formation is generated.
As a result, the positioning error of the proposed system becomes significantly smaller than that of a dead-reckoning system as travel time increases.
\end{itemize}

The remainder of this paper is organized as follows: the movement pattern of the robots maintaining the equilateral triangular formation is first introduced in Section~\ref{sec:equilateral}.
Next, the collision avoidance algorithm based on the proposed formation is described in Section~\ref{sec:path_planning}, followed by experimental results from partial movement steps using four mobile robots in Section~\ref{sec:localization}.
To demonstrate the performance of the proposed algorithm in wide open spaces, simulation results based on realistic error models derived from the experiments are presented in Section~\ref{sec:simulation}.
Finally, the conclusions of this study are provided in Section~\ref{sec:conclusions}.

\section{Proposed equilateral triangular formation for swarm robot localization}
\label{sec:equilateral}

Although we conceived the initial idea of triangular formation-based robot localization early on \citep{Kim17:Simulation}, implementing such a system without relying on distance-measuring sensors proved challenging.
If accurate distance-measuring sensors, such as laser range finders, are available, a moving robot can easily localize itself using distance information from two beacon robots at known locations.
This constitutes a straightforward trilateration problem.
However, our goal is to develop a highly accurate localization system based solely on low-cost monocular vision sensors and visual markers, in order to reduce system cost and ensure applicability to low-cost swarm robots.
Estimating distances to beacon robots through vision processing is not well-suited for our purpose, as it incurs high computational cost and offers relatively low accuracy compared to distance-measuring sensors.

\begin{figure}
  \centering
  \includegraphics[width=0.9\linewidth]{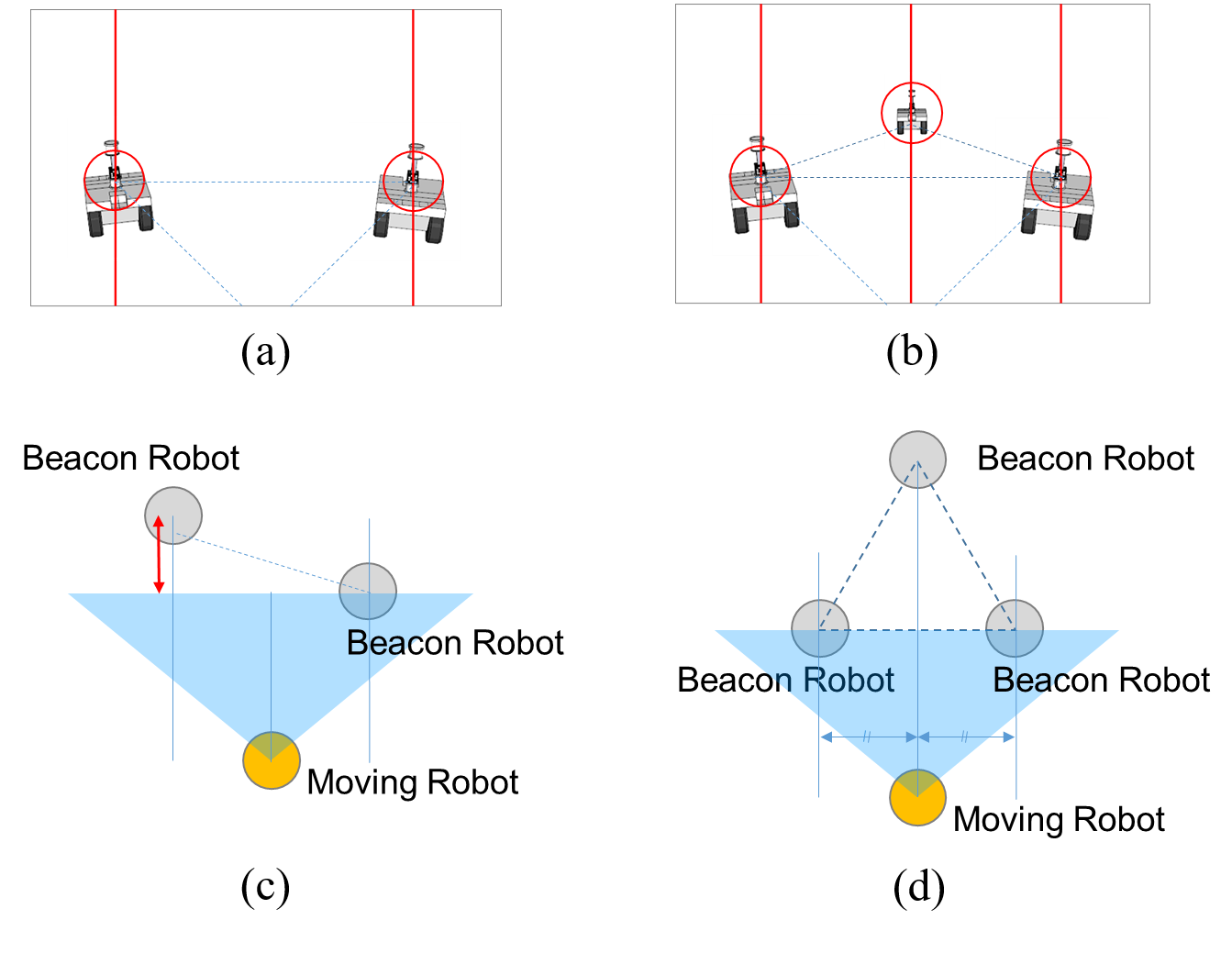}
  \caption{Triangular formation for localization with three robots vs. four robots.
    (a) View from the moving robot’s perspective corresponding to the three-robot formation shown in (c).  
    (b) View from the moving robot’s perspective corresponding to the four-robot formation shown in (d). 
    (c) Arbitrary triangular formation based on three robots. 
    (d) Equilateral triangular formation based on four robots.}
\label{fig:TriFormConcept}
\end{figure}

To maintain the position of each robot at the vertices of a triangle with high accuracy while marching toward the destination, we propose an equilateral triangular formation consisting of four robots.
This approach substantially simplifies the process of accurately positioning each robot at its designated vertex.
The moving robot determines its target vertex by checking the lateral distances to the three beacon robots, as shown in Fig.~\ref{fig:TriFormConcept}(d), and depth information is not required in our approach (i.e., a monocular camera is sufficient).

Accurate depth information is more difficult to obtain than accurate lateral distance information when using only a low-cost monocular vision sensor and visual markers.
In contrast, lateral distance information can be acquired with high accuracy through simple pixel counting in the vision image.
Moreover, the use of an equilateral triangular formation facilitates straightforward generalization of the method to systems with more than four robots (details are discussed in Section~\ref{sec:path_planning}).

Theoretically, an equilateral triangle can be formed using three robots.
In this case, however, the depth error indicated by the red arrow in Fig.~\ref{fig:TriFormConcept}(c) must be mitigated through vision processing.
The lateral distance information obtained via simple pixel counting cannot compensate for this error.
If depth information is not available, the moving robot may mistakenly assume that it is positioned at the right vertex of the equilateral triangle based on its vision image in Fig.~\ref{fig:TriFormConcept}(a).
Even when vision processing is applied to extract depth information, the resulting accuracy remains low when using only a monocular vision camera and visual markers, and the localization performance of the moving robot cannot be reliably maintained.

Therefore, we propose a four-robot system, as illustrated in Fig.~\ref{fig:TriFormConcept}(d).
The middle robot in the moving robot's field of view (FOV) (Fig.~\ref{fig:TriFormConcept}(b)) acts as an anchor, allowing the moving robot to reach the correct vertex location by checking only the lateral distances between the beacon robots.
Since depth information is not required in this approach, the localization accuracy of the moving robot can be higher than that of the three-robot system.
Given that our application targets swarm robots, which typically consist of more than four units, the use of one additional robot to improve localization accuracy is not a limitation of the proposed approach.

\begin{figure}
  \centering
  \includegraphics[width=0.6\linewidth]{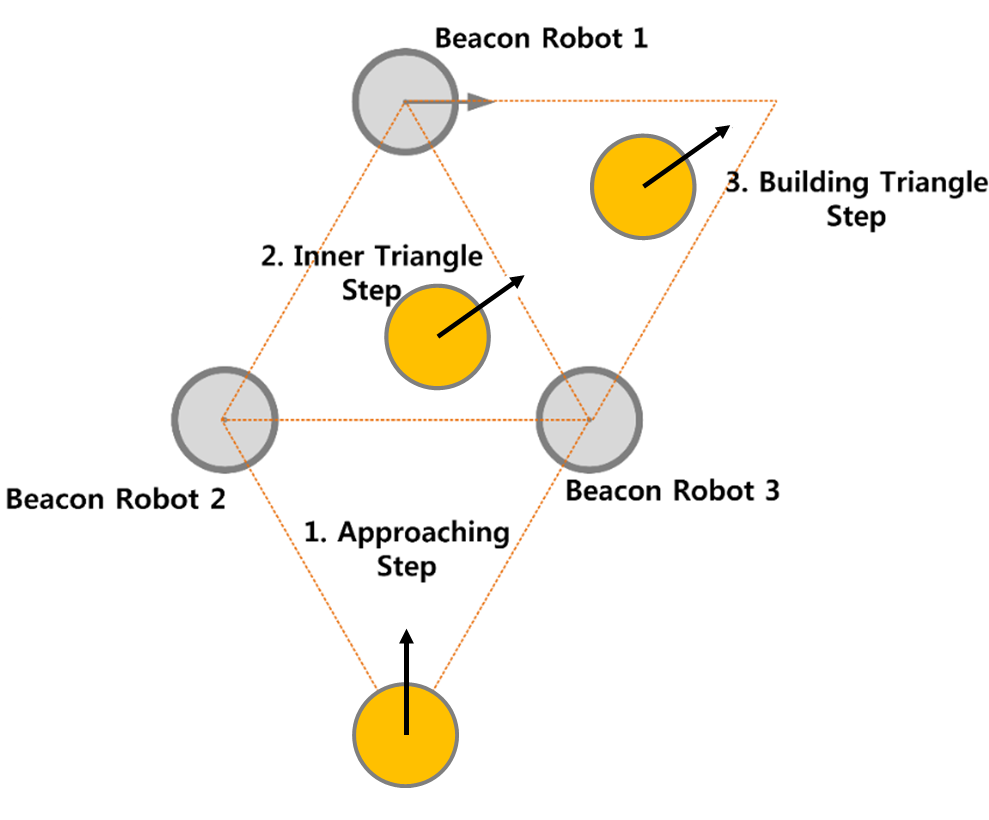}
  \caption{Movement of a robot to the target vertex to form the next equilateral triangular formation.
  The movement consists of three steps: the approaching step, the inner triangle step, and the building triangle step.}
\label{fig:MoveToTargetVertex}
\end{figure}

Fig.~\ref{fig:MoveToTargetVertex} illustrates the maneuver of the moving robot relative to three beacon robots, which remain temporarily stationary at the vertices of the equilateral triangle until the moving robot reaches the target vertex to form the next triangle.
The procedure for the moving robot to reach the target vertex consists of three steps: an approaching step, an inner triangle step, and a building triangle step.

In the approaching step, the moving robot advances toward the triangular formation of the beacon robots using its front-view monocular camera.
In the inner triangle step, the moving robot maneuvers toward the target vertex.
In the building triangle step, the moving robot locates and moves to the correct vertex position to form a new equilateral triangle using its rear-view monocular camera.
As shown in Fig.~\ref{fig:TriFormConcept}(b), the moving robot operates independently of odometry or ranging sensors when determining the correct vertex position.
Once the moving robot reaches the target vertex, its role changes to that of a beacon robot, one of the previous beacon robots becomes the new moving robot, and the formation continues marching toward the destination.

\begin{figure}
  \centering
  \includegraphics[width=0.60\linewidth]{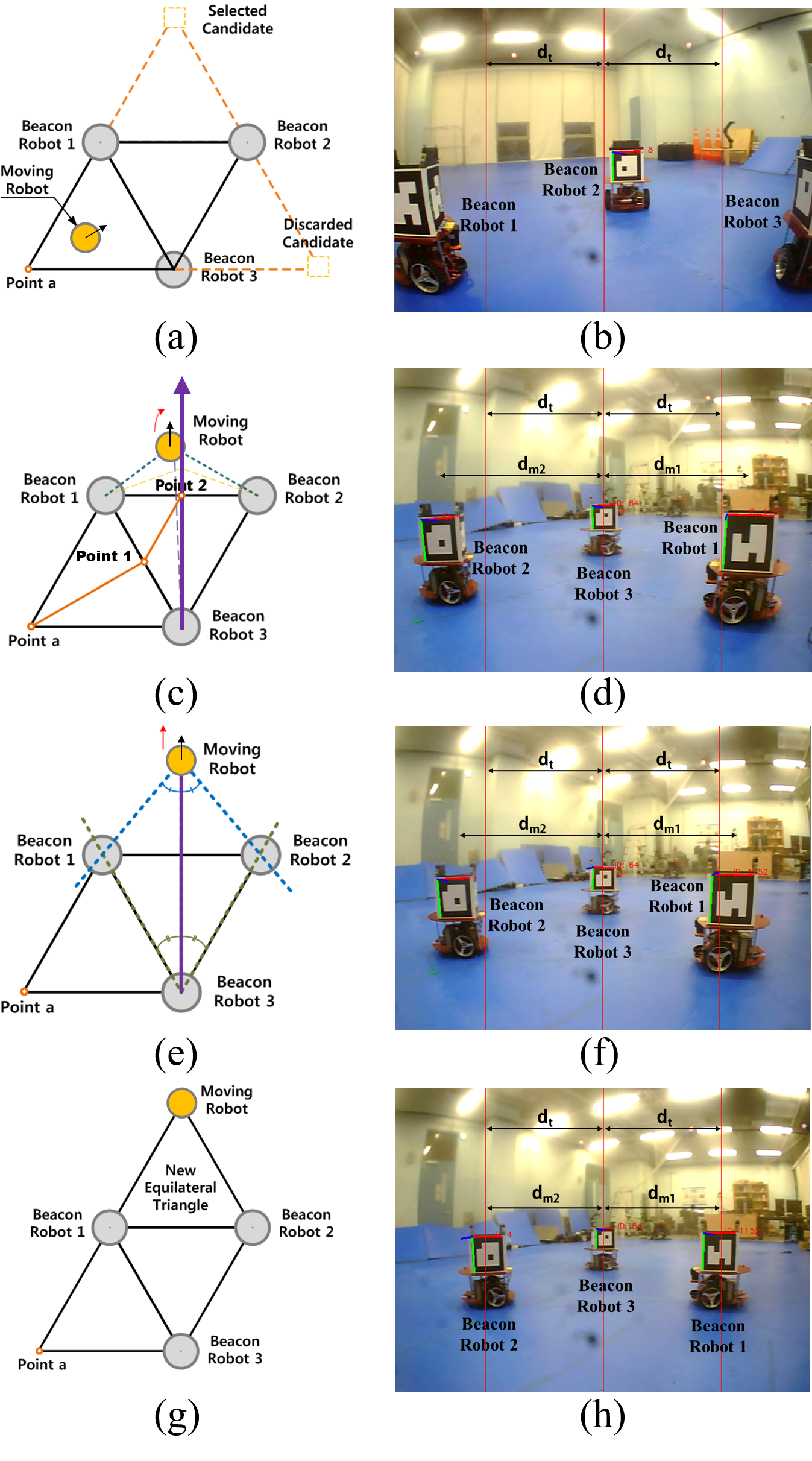}
  \caption{Algorithm used to create the next equilateral triangle in the desired moving direction.
    (a) The moving robot selects the target vertex and begins its maneuver.
    (b) Front-view image during the approaching step.
    (c) The orange line indicates the path of the moving robot. The purple arrow represents the extended height of the triangle.
    After passing point 2, the moving robot is controlled to maintain equal lateral distances to beacon robots 1 and 2 with respect to beacon robot 3.
    (d) The moving robot uses its rear-view camera to estimate the lateral distances between the beacon robots and is controlled to make d\textsubscript{m1} and d\textsubscript{m2} equal.
    (e) The moving robot moves forward until the lateral distances between beacon robot 3 and the two nearby beacon robots (i.e., distances d\textsubscript{m1} and d\textsubscript{m2}) match the pre-established value (i.e., d\textsubscript{t}).
    (f) The lateral distances to the two nearby beacon robots (i.e., d\textsubscript{m1} and d\textsubscript{m2}) are equal, but they do not yet match the desired value d\textsubscript{t}, which is determined by the size of the equilateral triangle.
    (g) The moving robot settles at the target vertex to complete the new equilateral triangular formation.
    (h) Rear-view image after the moving robot has settled at the vertex. At this point, d\textsubscript{m1}, d\textsubscript{m2}, and d\textsubscript{t} are all equal.}
\label{fig:TriFormAlgorithm}
\end{figure}

The detailed procedure for the moving robot to reach the target vertex in the building triangle step is presented in Algorithm~I and Fig.~\ref{fig:TriFormAlgorithm}.
After the moving robot passes between beacon robots 1 and 3, as shown in Fig.~\ref{fig:TriFormAlgorithm}(a), it continues moving between beacon robots 1 and 2, as shown in Fig.~\ref{fig:TriFormAlgorithm}(c).
Up to this point, the robot uses its front-view camera.
It then follows the steps in Algorithm~I using its rear-view camera.
In Fig.~\ref{fig:TriFormAlgorithm}(b), \ref{fig:TriFormAlgorithm}(d), \ref{fig:TriFormAlgorithm}(f), and \ref{fig:TriFormAlgorithm}(h), the center red vertical line indicates the robot’s center of sight.
The left and right red vertical lines are equidistant from the center line, with a spacing of d\textsubscript{t}.
Here, d\textsubscript{t} represents the desired lateral pixel distance between the two side markers in the robot’s field of view.
The value of d\textsubscript{t} is selected based on the constraints of the field-of-view angle of the monocular camera used in the experiment.
In this study, it was set to 280 pixels.
This parameter can be adjusted depending on the camera specifications and the desired scale of the swarm robot localization system.
This desired spacing is ultimately determined by the side length of the equilateral triangle.

\paragraph{Algorithm I}
\begin{enumerate}
    \item Maneuver the moving robot so that the beacon robot located at the opposite vertex (i.e., beacon robot 3) appears at the center of the moving robot’s rear-view image, as shown in Fig.~\ref{fig:TriFormAlgorithm}(c) and \ref{fig:TriFormAlgorithm}(d).
    
    \item Calculate d\textsubscript{m1} and d\textsubscript{m2}, the lateral distances between robots 3 and 1, and robots 3 and 2, respectively, using pixel counting.
    
    \item If d\textsubscript{m1} is greater than d\textsubscript{m2}, turn toward beacon robot 1; otherwise, turn toward beacon robot 2, to make d\textsubscript{m1} and d\textsubscript{m2} equal, as shown in Fig.~\ref{fig:TriFormAlgorithm}(e) and \ref{fig:TriFormAlgorithm}(f).
    
    \item Compare d\textsubscript{m1} and d\textsubscript{m2} to the given target disparity value, d\textsubscript{t}, which is determined by the side length of the equilateral triangle.
    
    \item If d\textsubscript{m1} and d\textsubscript{m2} are greater than d\textsubscript{t}, move the robot forward (in the direction of movement); otherwise, move it backward.
    
    \item When d\textsubscript{m1}, d\textsubscript{m2}, and d\textsubscript{t} become equal, stop the moving robot at that position. At this point, a new equilateral triangular formation is completed, as shown in Fig.~\ref{fig:TriFormAlgorithm}(g) and \ref{fig:TriFormAlgorithm}(h).
\end{enumerate}

Note that the depths of the robots in the image are not estimated through vision processing due to their relatively low estimation accuracy. 
Instead, only lateral distances in the image are utilized in Algorithm~I, which provides significantly higher accuracy.
Our method places the moving robot at the desired two-dimensional location based solely on one-dimensional lateral distance information, which can be extracted with high accuracy even using a low-cost monocular vision sensor and minimal computational power.

\section{Path planning algorithm for the proposed localization method}
\label{sec:path_planning}

\begin{figure}
  \centering
  \includegraphics[width=0.7\linewidth]{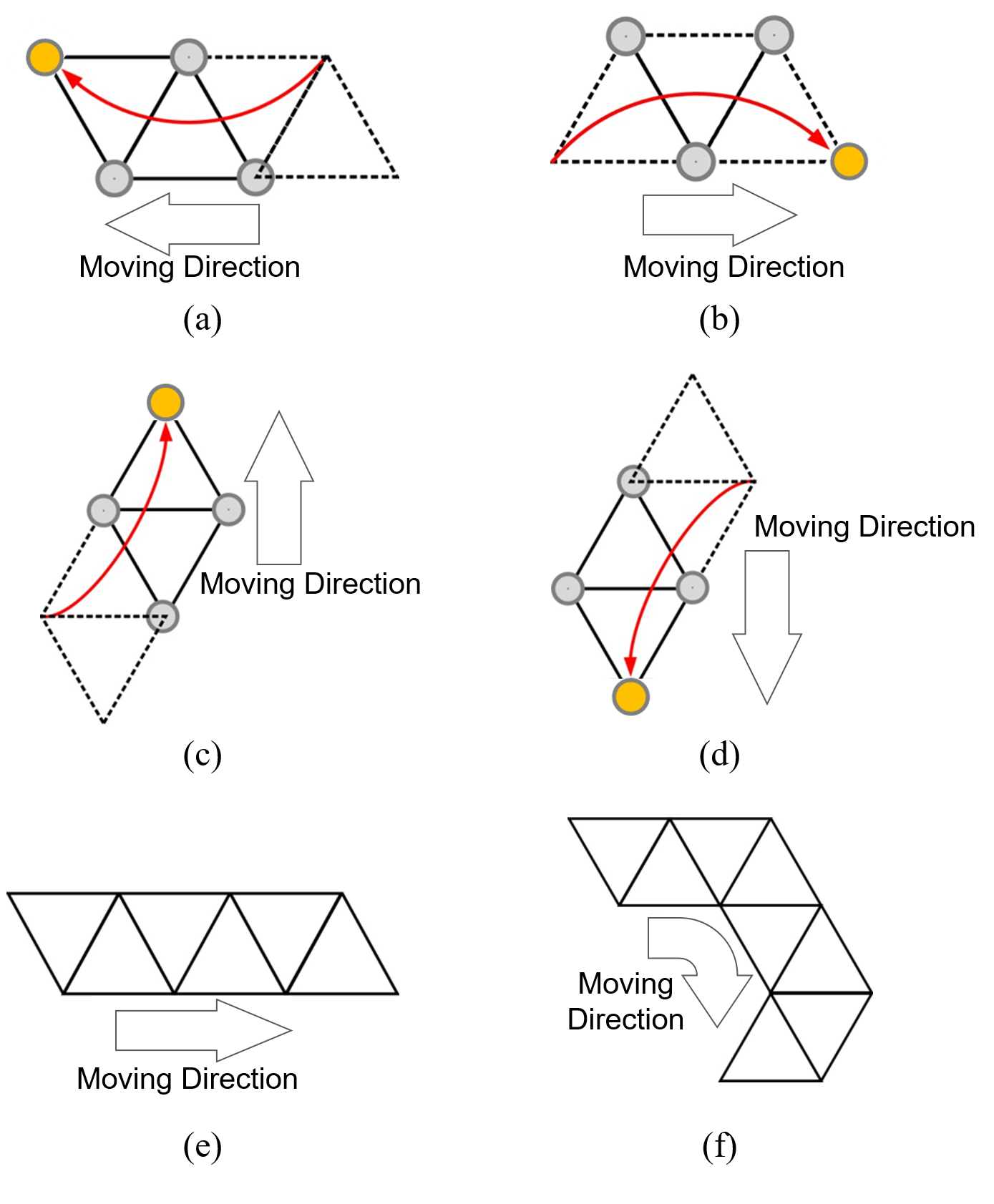}
  \caption{Movement patterns of the triangular formation.
    (a) A single leftward movement step.
    (b) A single rightward movement step.
    (c) A single upward movement step.
    (d) A single downward movement step.
    (e) Straight-line movement through multiple rightward steps.
    (f) Turning motion achieved through a sequence of rightward, downward, and leftward steps.}
\label{fig:MovingPattern}
\end{figure}

The proposed system selects one moving robot and its local destination (i.e., target vertex) in each step to advance the entire formation toward the destination.
The triangular formation can move in four directions by selecting an appropriate robot within the formation as the moving robot.
Once the desired moving direction is determined, the corresponding moving robot is selected.
For example, if the formation moves left or upward, the robot located to the right or below is chosen as the moving robot, as shown in Fig.~\ref{fig:MovingPattern}(a) and \ref{fig:MovingPattern}(c), respectively, while the others serve as beacon robots.
Fig.~\ref{fig:MovingPattern} illustrates the various movement patterns of the triangular formation.
In Fig.~\ref{fig:MovingPattern}(a)--\ref{fig:MovingPattern}(d), the grey circles represent beacon robots and the orange circle represents the moving robot.
Fig.~\ref{fig:MovingPattern}(e) and \ref{fig:MovingPattern}(f) show combined movements composed of the basic steps shown in Fig.~\ref{fig:MovingPattern}(a)--\ref{fig:MovingPattern}(d).
As demonstrated in Fig.~\ref{fig:MovingPattern}, the triangular formation is capable of moving in any direction.

\begin{figure}
  \centering
  \includegraphics[width=0.6\linewidth]{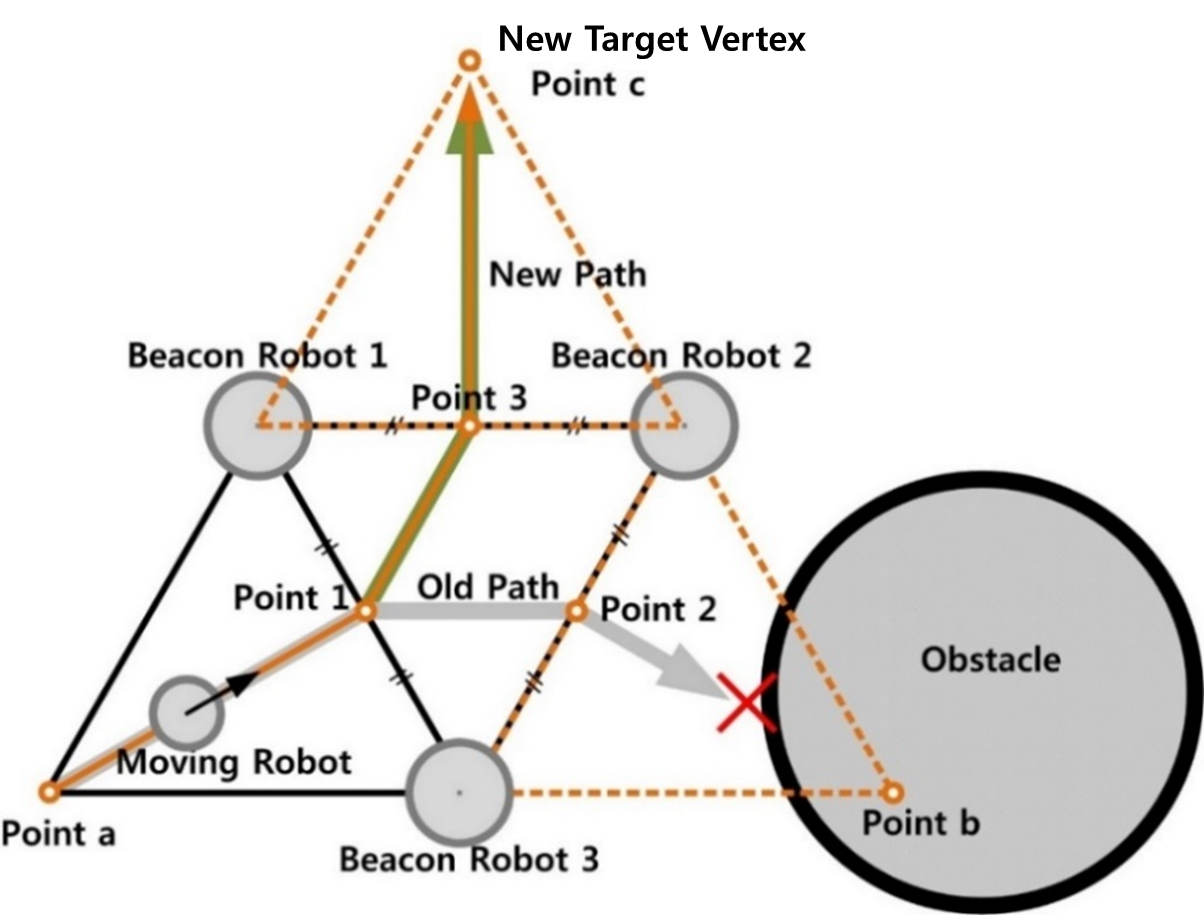}
  \caption{Path for collision avoidance with an obstacle in a candidate path.
  Both paths P1=\{point 1, point 3, point c\} and P2=\{point 1, point 2, point b\} can avoid inner collisions, but only path P1 avoids outer collisions.}
\label{fig:CollAvoid}
\end{figure}

During the movement of the formation, collisions between robots and obstacles must be avoided. 
Collisions are categorized into two types: \textit{inner collisions}, which occur between robots within the formation, and \textit{outer collisions}, which occur between robots and external obstacles in the environment. 
First, inner collisions are avoided using Algorithm~II. 
The set P = \{point 1, point 3, point c\}, obtained by Algorithm~II, correctly represents the desired path for the moving robot, as shown in Fig.~\ref{fig:CollAvoid}.

\paragraph{Algorithm II}
\begin{enumerate}
    \item Construct a set of center points C between the beacon robots (i.e., C = \{point 1, point 2, point 3\} in Fig~\ref{fig:CollAvoid}).
    
    \item Include in set P the point in C that is closest to the moving robot (i.e., point 1 in Fig.~\ref{fig:CollAvoid}), which serves as the first point in the desired path.
    
    \item Include in set P the point in C that is closest to the target vertex (i.e., point 3 in Fig.~\ref{fig:CollAvoid} when the target vertex is point c).
    
    \item Include the target vertex itself (i.e., point c in Fig.~\ref{fig:CollAvoid}) in set P.
\end{enumerate}

After inner collision avoidance is achieved, the robot formation must also avoid external obstacles (i.e., outer collision avoidance).
To this end, we assume that the robots can detect obstacles using low-cost detection sensors, such as ultrasonic sensors \citep{Rhee18:Ground, Rhee19:Low}, and that obstacle information is shared among the robots via any communication channel (e.g., WiFi in our testbed).
Since the obstacle locations are not used for localization purposes in our system, high-cost distance-measuring sensors are unnecessary, and approximate information about the presence of obstacles is sufficient.

Let us assume that point b in Fig.~\ref{fig:CollAvoid} is the closest vertex to the final destination of the robot formation.
In this case, it would be reasonable for the moving robot to proceed toward point b.
However, since there is an obstacle along the path to point b in Fig.~\ref{fig:CollAvoid}, the robot should instead select a different vertex to avoid a collision.
If the next closest vertex to the destination is point c, the robot selects point c as its new target.
In summary, the priority of candidate vertices in our algorithm is determined by their Euclidean distance to the destination.

\begin{figure}
  \centering
  \includegraphics[width=0.6\linewidth]{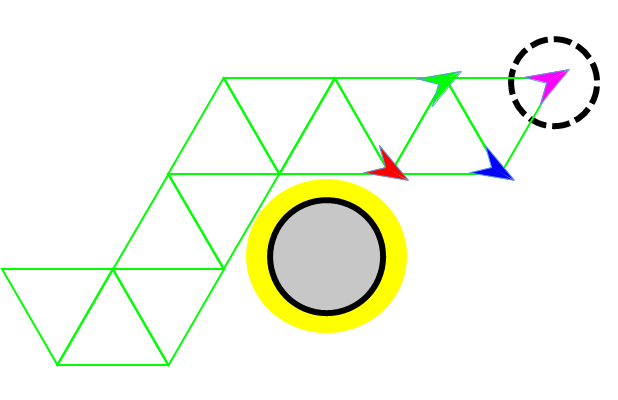}
  \caption{Example path of the robot formation to the destination.
    The yellow region around the obstacle represents the safety zone, and the formation detours around this zone to avoid potential collisions.}
\label{fig:PathToDest}
\end{figure}

Fig.~\ref{fig:PathToDest} shows an example in which the robot formation detours around an obstacle and reaches the destination using the ``closest vertex'' algorithm.
The yellow circle in Fig.~\ref{fig:PathToDest} represents the safety zone.
If the path to an initially selected vertex intersects this safety zone, the moving robot should instead aim for the next closest vertex to ensure safety.
If the robot formation is in a more complex situation---where obstacles are located near both points b and c in Fig.~\ref{fig:CollAvoid}---the system should select a different robot as the moving robot and choose the next closest vertex, other than points b and c, as the new target.
While the current implementation effectively handles simple obstacle avoidance by selecting the next closest vertex, its performance in more complex environments with multiple closely spaced obstacles (e.g., near both points b and c in Fig.~\ref{fig:CollAvoid}) depends on the ability to dynamically reassign the moving robot and replan the target vertex.
For even more intricate scenarios, the integration of higher-level global path planning algorithms, such as A* or rapidly exploring random tree (RRT), could further enhance the system's ability to navigate safely and efficiently through cluttered environments.

\begin{figure}
  \centering
  \includegraphics[width=0.8\linewidth]{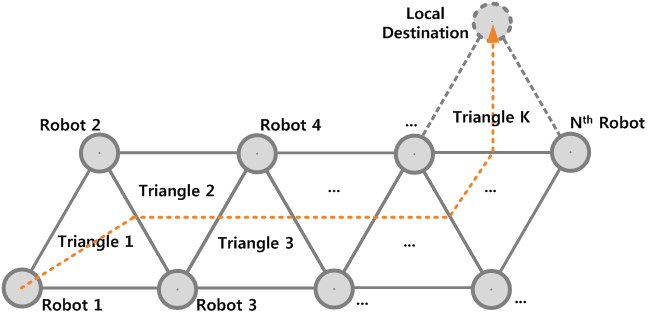}
  \caption{Movement of an $N$-robot system.
    This configuration is a direct extension of the proposed four-robot system.}
\label{fig:NrobotSystem}
\end{figure}

The proposed robot formation can be readily extended to an $N$-robot system, as illustrated in Fig.~\ref{fig:NrobotSystem}.
In such a system, each robot participates in at least one equilateral triangular formation.
In the basic implementation, the farthest robot from the destination is selected as the moving robot, and it moves to the target vertex (i.e., local destination), which is the closest possible vertex to the global destination.
This algorithm generates a ``snake-like'' movement of the formation toward the destination and is applicable to swarm robot systems \citep{Kim16:Mapping}.

\section{Localization performance and discussion}
\label{sec:localization}

The proposed localization method, based on equilateral triangular formations, was tested using four mobile robots.
Localization error models obtained in a controlled laboratory environment, as explained in Section~\ref{Sec:ErrorModels}, were then used to perform realistic performance simulations in larger open spaces.

\subsection{Experimental setup}

\begin{figure}
  \centering
  \includegraphics[width=0.7\linewidth]{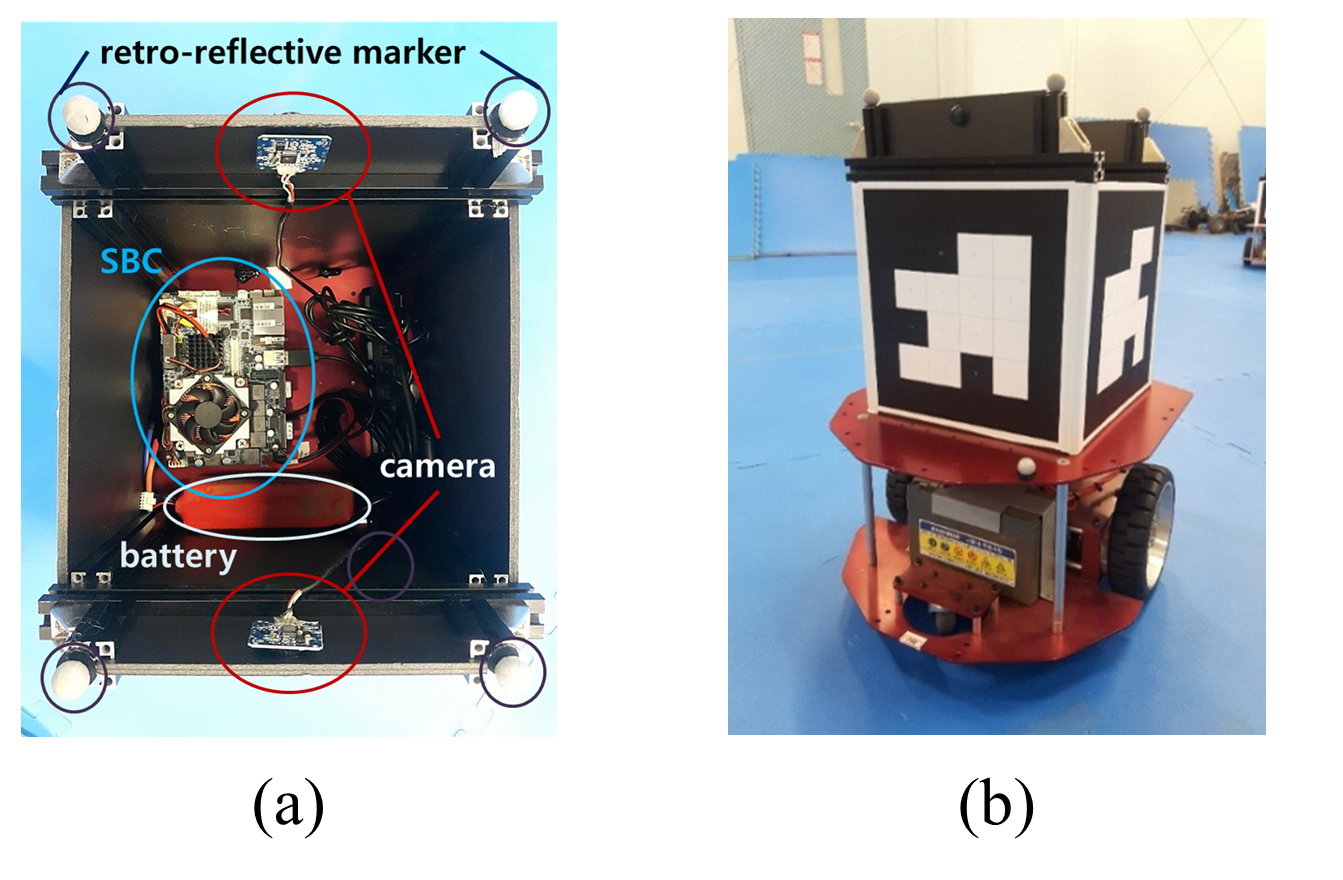}
  \caption{Mobile robot platform used for the experiments.
    (a) System components. Two cameras are mounted to view the front and back, and a single board computer (SBC) controls the entire system.
    Retro-reflective markers are attached to enable ground truth motion capture using a commercial system.
    (b) Exterior of the mobile robot. Visual markers are placed on all sides.}
\label{fig:RobotPlatform}
\end{figure}

A commercial mobile robot named Stella (Fig.~\ref{fig:RobotPlatform}) was used for the experiments.
Each robot was equipped with a single board computer (SBC), two cameras, and four visual markers.
The SBC handled basic image processing and pixel counting to maintain the robot formation, as well as steering and velocity control.
One camera faced forward and the other faced backward.
In general, two additional cameras can be mounted on the left and right sides of the robot to enable omnidirectional exploration.

Visual markers were installed on all four sides of the robots.
Using simple image processing based on these visual markers, the moving robot extracts the center point of each beacon robot.
The AprilTag marker detection system \citep{Olson2011:AprilTag} was employed to determine the position of each robot.
Once the center points are identified, accurate lateral distances between the robots are obtained through simple pixel counting.
A WiFi communication module in the SBC was used to share path planning information, including obstacle data, among the robots.
It is not necessary for the robots to share any visual images for localization purposes; thus, the required communication bandwidth of the system remains very low.

\begin{figure}
  \centering
  \includegraphics[width=0.6\linewidth]{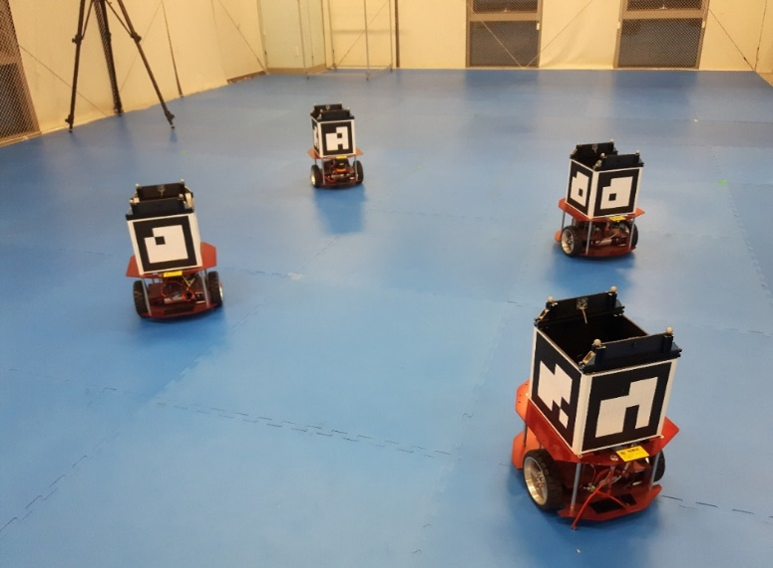}
  \caption{Four-robot system in a controlled laboratory environment with a commercial motion capture system.
   Ten motion capture cameras provided ground truth data for the movement of each robot during the experiment.}
\label{fig:FourRobots}
\end{figure}

In order to evaluate the localization errors, ground truth data of the robots' actual movements must be obtained.
To this end, a 10-camera Vicon MX motion capture system (Vicon Motion Systems Ltd., Oxford, UK) was used.
This system provides the positions of retro-reflective markers attached to the robots, as shown in Fig.~\ref{fig:RobotPlatform}(a), with sub-centimeter accuracy.
The accuracy for stationary markers can be further improved by averaging the measured positions over a longer duration.
The experimental setup with four robots in the motion capture studio is shown in Fig.~\ref{fig:FourRobots}.
This controlled laboratory environment enables accurate evaluation of localization errors, although the space covered by the ten cameras is limited.
Accordingly, realistic error models of the proposed system were obtained in this environment, and performance in larger open spaces was evaluated through simulations based on these models.

\subsection{Error models}
\label{Sec:ErrorModels}

\begin{figure}
  \centering
  \includegraphics[width=0.6\linewidth]{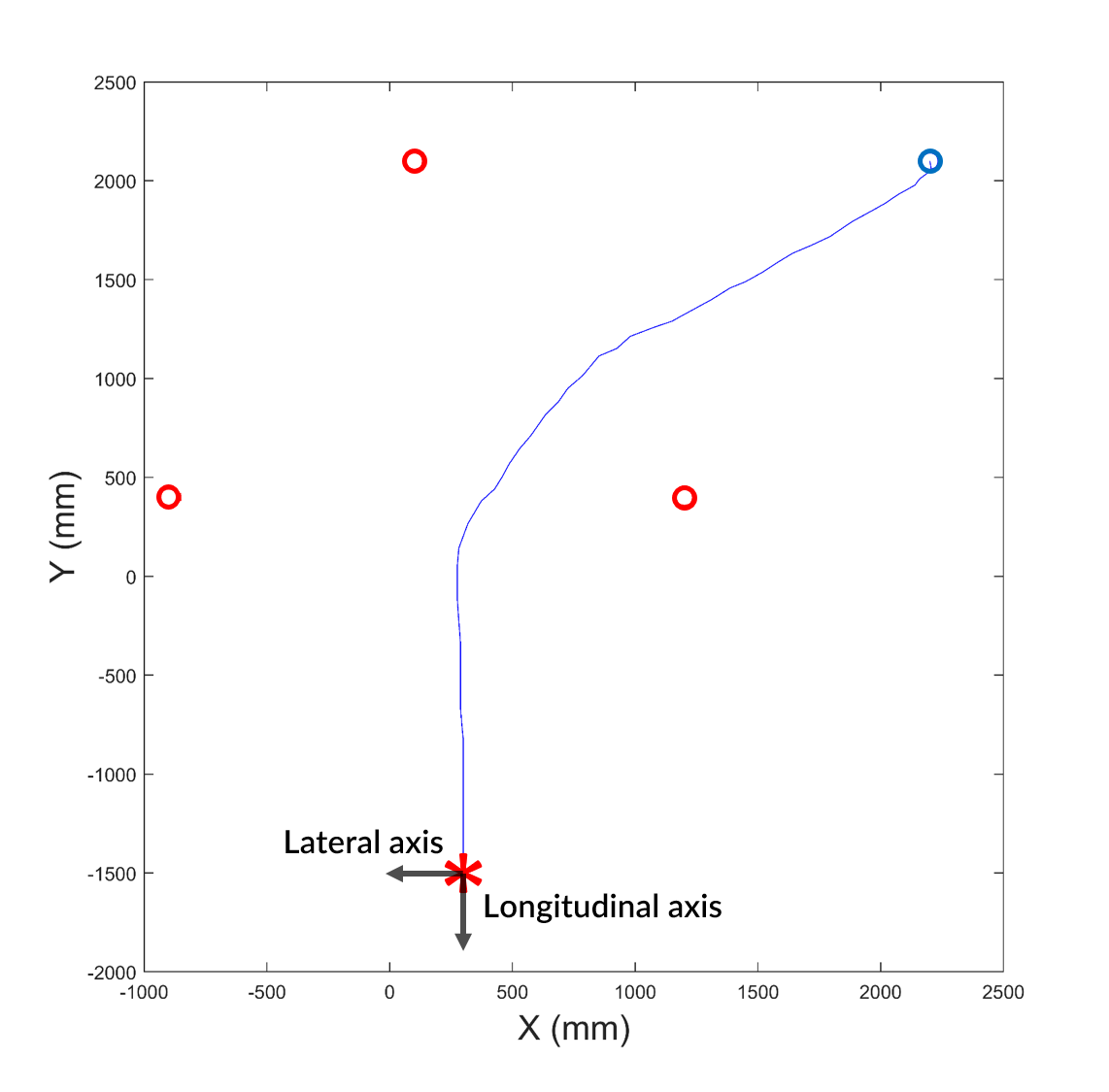}
  \caption{Example trajectory of a moving robot passing through the previous triangular formation of three beacon robots to create a new triangular formation.
   The robot moved from the blue circle to the red asterisk, which represents a vertex of the new equilateral triangle.
   This ground truth trajectory was obtained using a commercial motion capture system in a controlled laboratory environment.}
\label{fig:TrajExample}
\end{figure}

During the experiments, the moving robot approached the target vertex by following the steps described in Algorithm~I and illustrated in Fig.~\ref{fig:TriFormAlgorithm}.
An example trajectory of the moving robot while forming a new equilateral triangle is shown in Fig.~\ref{fig:TrajExample}.
The red circles in Fig.~\ref{fig:TrajExample} represent the positions of the three beacon robots.
The blue circle and the red asterisk indicate the initial position and the local destination (i.e., the target vertex) of the moving robot, respectively.
The blue line shows the trajectory taken by the moving robot.

\begin{figure}
  \centering
  \includegraphics[width=0.6\linewidth]{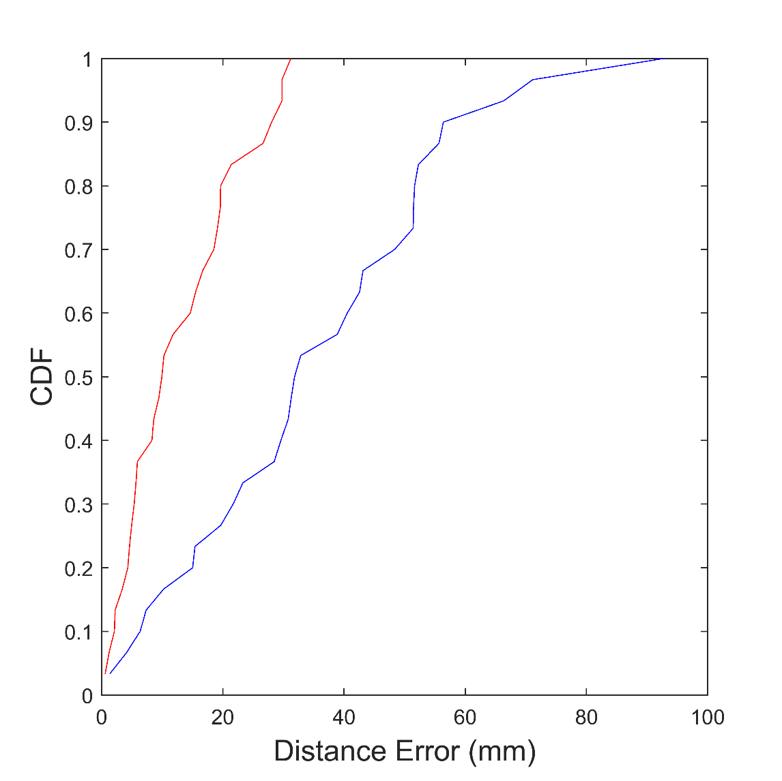}
  \caption{Empirical cumulative distribution functions (CDFs) of the distance errors for the proposed system along the lateral axis (blue) and the longitudinal axis (red).}
\label{fig:CDF}
\end{figure}

The experiment in which a new equilateral triangle is formed was repeated 60 times in a controlled laboratory environment, and statistics of the localization error were collected.
The empirical cumulative distribution functions (CDFs) of the localization errors along the lateral and longitudinal axes, based on the 60 trials, are shown in Fig.~\ref{fig:CDF}.
The lateral axis is defined as parallel to the base of the new triangle, and the longitudinal axis as perpendicular to the base, as illustrated in Fig.~\ref{fig:TrajExample}.

The mean localization error along the lateral and longitudinal axes was calculated as
\begin{equation}
    \mu_x = \frac{1}{N} \sum_{i=1}^{N} e_{x,i}, \quad \mu_y = \frac{1}{N} \sum_{i=1}^{N} e_{y,i}
\end{equation}
where $e_{x,i}$ and $e_{y,i}$ represent the lateral and longitudinal error samples, respectively, and $N$ is the total number of measurements.
The corresponding standard deviations were computed as
\begin{equation}
    \sigma_x = \sqrt{\frac{1}{N-1} \sum_{i=1}^{N} (e_{x,i} - \mu_x)^2}, \quad \sigma_y = \sqrt{\frac{1}{N-1} \sum_{i=1}^{N} (e_{y,i} - \mu_y)^2}
\end{equation}

The mean localization error along the lateral and longitudinal axes was 36~mm and 13~mm, respectively.
The standard deviation of the localization error along the lateral and longitudinal axes was 21~mm and 9~mm, respectively.
The lateral axis error is generally greater than the longitudinal axis error because the robot used in the experiment has a nonholonomic constraint.
This constraint makes it difficult for the robot to move precisely in the lateral direction.

The performance of the proposed method was compared with that of a conventional dead-reckoning method, which represents a typical low-cost localization approach.
This provides a fair comparison between two low-cost localization methods: the proposed method and the dead-reckoning method.

The error models of conventional dead-reckoning systems have been studied in \citep{Gebre-Egziabher14} and \citep{Vezinet14}.
A widely used approach to implementing a low-cost dead-reckoning system involves measuring the heading angle using a compass or magnetometer and estimating speed using an odometer or speed sensor.
Microelectromechanical systems (MEMS) gyros or accelerometers can also be used to estimate the heading angle and speed.
Thus, the performance of such a dead-reckoning system depends on the estimation errors of heading and speed derived from the applied sensors.

In \citep{Gebre-Egziabher14}, the heading angle error was modeled as a Gauss-Markov process with a correlation time of 120 seconds and a standard deviation of 2.5 degrees.
These values were obtained based on actual measurements of estimation errors from a low-cost inertial navigation system (i.e., MEMS gyros and accelerometers) and a set of magnetometers.
An additive white noise component with a standard deviation of 0.8 degrees was also included in the Gauss-Markov model.

The measurement model of a wheel speed sensor (WSS), which is typically used for dead reckoning, was proposed by \citep{Vezinet14} as follows: 
\begin{eqnarray}
\label{eqn:WheelSpeed}
	v_\mathrm{WSS} = (R + \varepsilon_R) (1 + S\!F_\mathrm{WSS}) \Omega_\mathrm{wheel} + n_\mathrm{WSS} 
\end{eqnarray}
where $v_\mathrm{WSS}$ is the linear velocity of the vehicle measured by the wheel speed sensor, $R$ is the nominal radius of the wheel, $\varepsilon_R$ is the bias on the nominal wheel radius, $S\!F_\mathrm{WSS}$ is a scale factor, $\Omega_\mathrm{wheel}$ is the intended wheel rotational speed, and $n_\mathrm{WSS}$ is the white Gaussian measurement noise.

The parameters of this model for our mobile robot platform, shown in Fig.~\ref{fig:RobotPlatform}, were experimentally estimated.
The obtained values were $R = 0.148$~m, $S\!F_\mathrm{WSS} = 0.16$, and the standard deviation of $n_\mathrm{WSS}$ was 0.045~m/s.
The $\varepsilon_R$ value is negligible because our mobile robot platform uses relatively small and rigid wheels.

We utilized the heading angle error model and the wheel speed sensor measurement model developed in \citep{Gebre-Egziabher14} and \citep{Vezinet14} to simulate the performance of a typical dead-reckoning system.
The performance of the simulated dead-reckoning system was subsequently compared with that of our proposed localization system, as presented in Section~\ref{sec:simulation}.

\subsection{Discussion}

Although camera distortion---particularly non-radial or asymmetric distortion---can affect the accuracy of marker detection, the symmetric arrangement and precise initial positioning of the beacon robots ensured that the initial triangular formation was accurately established.
As a result, the system maintained good localization accuracy without requiring additional dewarping or sub-pixel marker detection algorithms to compensate for center detection uncertainties.
For longer trajectories or more distortion-sensitive applications, integrating sub-pixel detection techniques could further enhance precision.

The parameters determining the size of the equilateral triangle formation---specifically d\textsubscript{m1}, d\textsubscript{m2}, and d\textsubscript{t}---were selected based on practical constraints imposed by the field-of-view angle of the commonly available monocular camera used in the experiments.
These values were chosen to ensure that all beacon robots remain visible within the camera's viewing angle during operation.
The triangle size directly influences controllability and planning flexibility: a smaller triangle can offer better maneuverability, especially in tight or cluttered environments, while a larger triangle may reduce the number of formation steps required to reach a destination, thereby improving mission efficiency.
The system is designed to be adaptable, allowing the triangle size to be tuned according to the camera’s viewing angle, resolution, and optical characteristics, as well as the performance requirements of the task.
This adaptability makes the system suitable for a wide range of cost-sensitive and mission-specific deployment scenarios.
It also highlights the potential for identifying an optimal triangle size tailored to specific mission profiles, balancing localization accuracy, path efficiency, and hardware limitations.

The parameters d\textsubscript{m1}, d\textsubscript{m2}, and d\textsubscript{t}, when properly matched in pixel units, help preserve the equilateral triangle geometry, resulting in localization errors that accumulate at a sub-pixel level during successive robot movements.
As demonstrated in our experiments, the proposed low-cost localization method achieves sufficiently high localization accuracy using a commercial monocular camera and a simple marker detection algorithm.
In scenarios involving longer trajectories without periodic error correction, higher localization performance can be achieved by employing more advanced hardware or vision systems.

\section{Simulation results}
\label{sec:simulation}

Based on the realistic error models of the proposed system---obtained in a controlled laboratory environment using the motion capture system---the performance of the proposed localization system in wide open spaces was evaluated through simulations.
As the proposed system is a low-cost solution that does not rely on accurate distance-measuring sensors or high computational power, it is appropriate to compare its performance with that of another low-cost solution, namely a dead-reckoning system.
The proposed system uses only monocular cameras, while the dead-reckoning system relies on wheel speed sensors and MEMS gyros and accelerometers.

\begin{figure}
  \centering
  \includegraphics[width=0.9\linewidth]{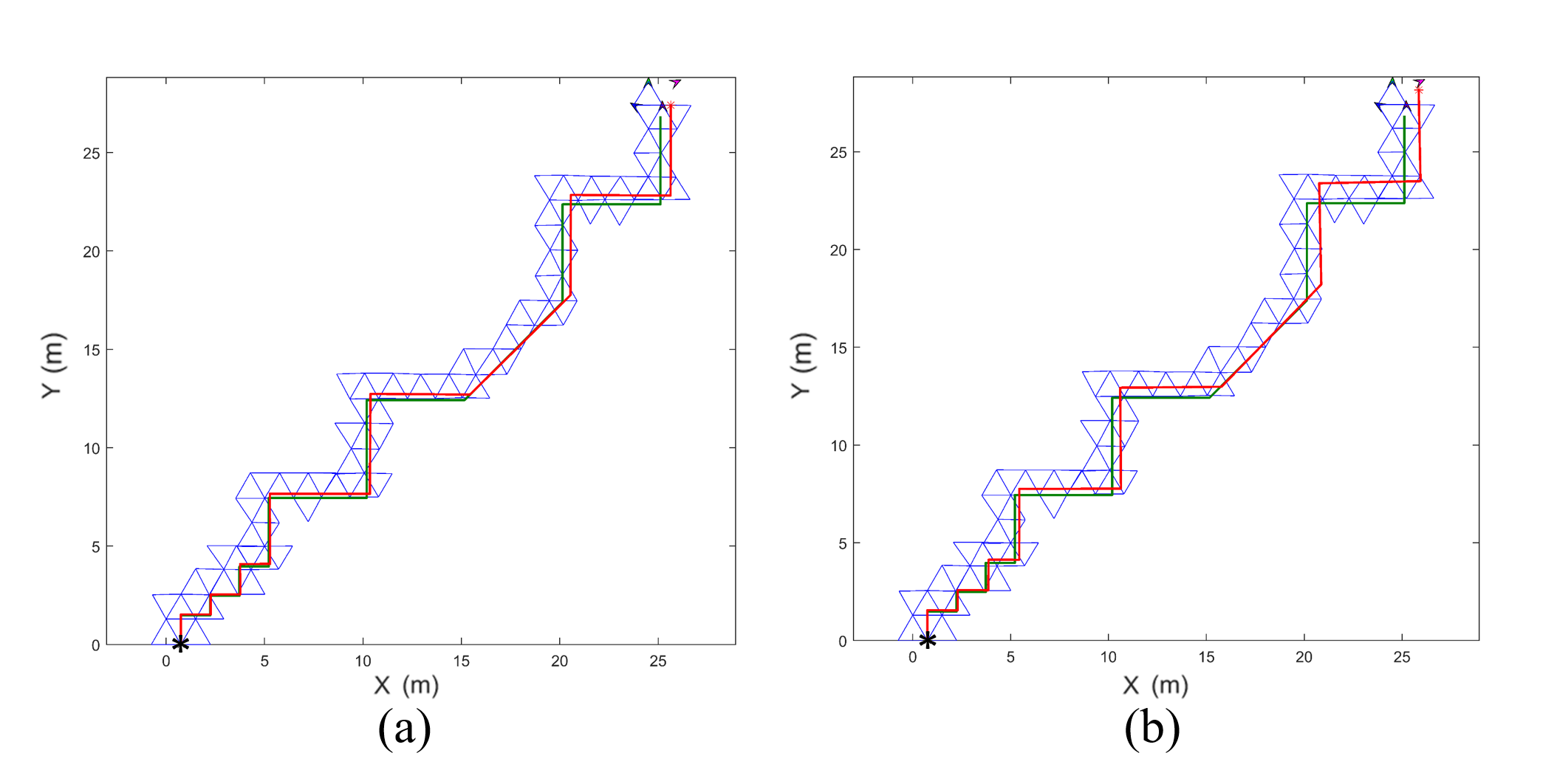}
  \caption{Comparison of the movement paths generated by the dead-reckoning method (red) and the formations of the proposed method (blue), following the given trajectory (green) with many turns.
  The $\Omega_\mathrm{wheel}$ in Eq.~(\ref{eqn:WheelSpeed}) was set to (a) 5.8~rad/s and (b) 2.9~rad/s.
  The localization error of the dead-reckoning method significantly increased when $\Omega_\mathrm{wheel}$ was reduced.}
\label{fig:SimTrajTurns}
\end{figure}

\begin{figure}
  \centering
  \includegraphics[width=0.9\linewidth]{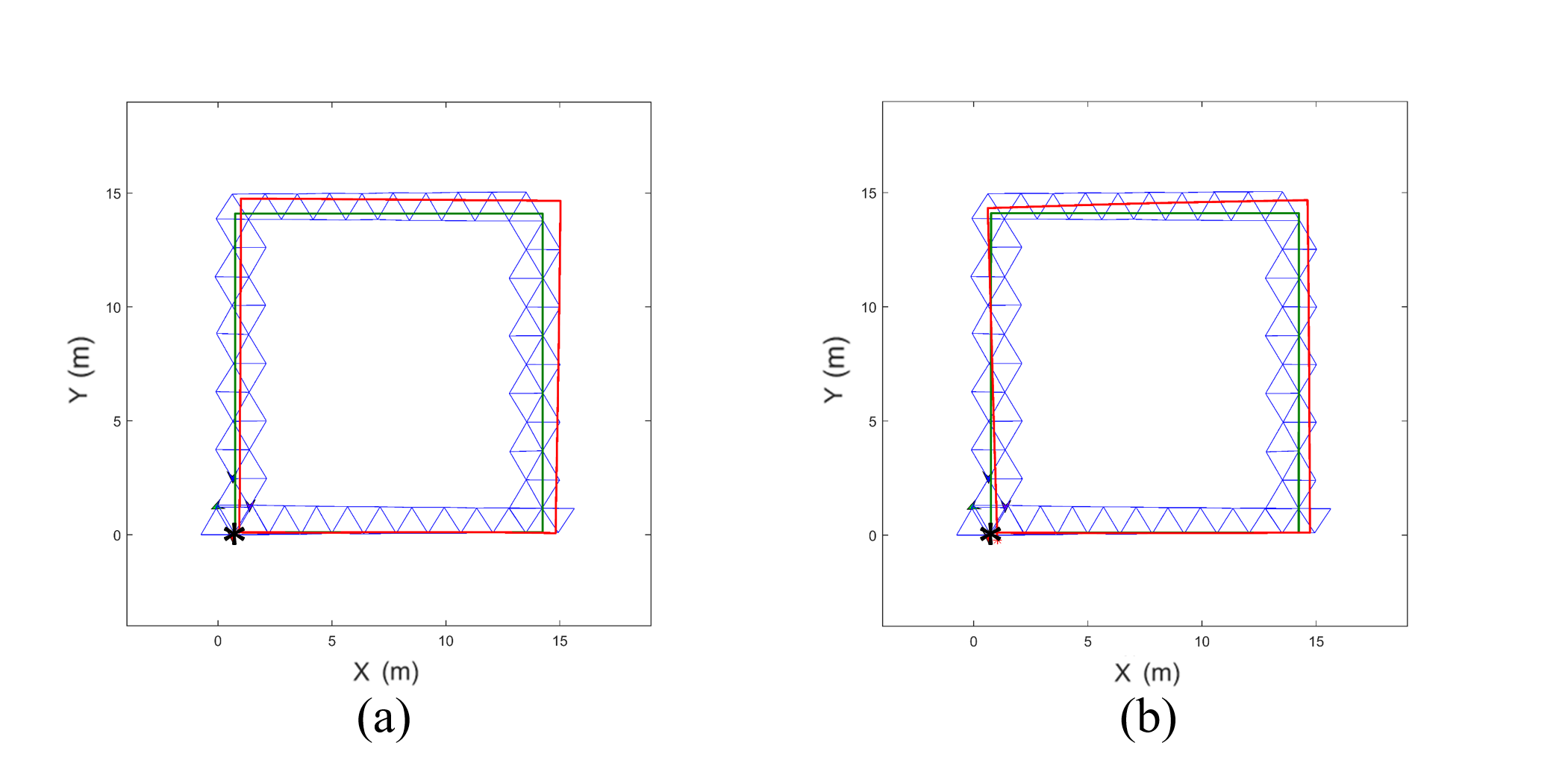}
   \caption{Comparison of the movement paths generated by the dead-reckoning method (red) and the formations of the proposed method (blue), following the given trajectory (green) representing a robot patrol case.
   The $\Omega_\mathrm{wheel}$ in Eq.~(\ref{eqn:WheelSpeed}) was set to (a) 5.8~rad/s and (b) 2.9~rad/s.
   The localization error of the dead-reckoning method significantly increased when $\Omega_\mathrm{wheel}$ was reduced.}
\label{fig:SimPatrol}
\end{figure}

The basic four-robot formation was evaluated through simulation.
The simulations were conducted along two trajectories: one with many turns (Fig.~\ref{fig:SimTrajTurns}) and another representing a robot patrol path (Fig.~\ref{fig:SimPatrol}).
The initial positions of the robots were set as (0.75~m, 0), (0, 1.30~m), (2.25~m, 0), and (1.5~m, 1.30~m).
The length of each side of the equilateral triangle was 1.5~m, and the initial heading angles of all robots were set to zero (i.e., facing east).

When the destination of the formation was specified, the farthest robot from the destination was selected as the moving robot, and the others served as beacon robots.
The selection of the moving robot is similar to the leader selection process in multi-robot systems, as described in Kwon \textit{et al.} \citep{Kwon15:Multiple}.
They proposed the multiple leader candidate (MLC) structure, in which the leader is competitively selected from among leader candidates (LCs) and guides the formation along the planned trajectory.
If the current leader experiences a fault or failure, the remaining LCs immediately compete to select a new leader, allowing the followers to maintain formation stability by tracking their vertices relative to the nearest LC until a new leader is designated.
The MLC structure offers scalability advantages, as LCs share information through a virtual-leader framework, enabling the system to accommodate additional robots by adjusting the number of LCs according to the capabilities of the communication devices.

While the goal is to form an ideal equilateral triangle geometry at each robot’s movement step, in practice, each robot reaches a slightly erroneous position, as characterized by the experimentally derived error model presented in Fig.~\ref{fig:CDF}.
This model shows the empirical error distributions along the lateral and longitudinal axes.
These position errors accumulate with each successive movement, and this error characteristic is taken into account in the simulation.

The green lines in Fig.~\ref{fig:SimTrajTurns} and Fig.~\ref{fig:SimPatrol} indicate the given trajectories, and the black asterisk marks the starting point. 
The proposed formations of robots from the starting point to the destination are shown in blue, while the path of a robot using the dead-reckoning system is shown in red in both figures.
The intended wheel rotational speed of the robots, i.e., $\Omega_\mathrm{wheel}$ in Eq.~(\ref{eqn:WheelSpeed}), was set to 5.8~rad/s for Fig.~\ref{fig:SimTrajTurns}(a) and 2.9~rad/s for Fig.~\ref{fig:SimTrajTurns}(b).

The localization errors of the dead-reckoning and proposed methods along the given trajectories in Fig.~\ref{fig:SimTrajTurns} and Fig.~\ref{fig:SimPatrol} are summarized in Table~\ref{table:CompLocErr}.
These errors represent average values based on 100 simulation runs. 
Since a vertex of a triangle cannot be placed arbitrarily, the localization error of the proposed system is defined as the difference between the actual location of the moving robot when it arrives at the final target vertex near the destination and the ideal vertex location, assuming all equilateral triangles are perfectly constructed from the starting point to the destination.

When the intended wheel rotational speed of the robots was reduced from 5.8~rad/s to 2.9~rad/s, the localization errors of the dead-reckoning system along the trajectories in Fig.~\ref{fig:SimTrajTurns} and Fig.~\ref{fig:SimPatrol} increased significantly.
In contrast, the localization errors of the proposed system did not change meaningfully.

\begin{table}
\centering
\caption{Comparison of localization errors along the trajectories in Fig.~\ref{fig:SimTrajTurns} and Fig.~\ref{fig:SimPatrol} with respect to the intended wheel rotational speed.}
\begin{tabular}{|c|c+c|c|} 
\hline
& $\Omega_\mathrm{wheel}$  & \multicolumn{2}{c|}{Localization error at the destination (m)}      \\ 
\cline{3-4}
& (rad/s) & Dead-reckoning & Proposed method   \\ 
\thickhline
\multirow{2}{*}{Trajectory in Fig.~\ref{fig:SimTrajTurns}} & 5.8 & 0.81 & 0.56      \\ 
\cline{2-4}
& 2.9 & 1.43 & 0.51   \\ 
\thickhline
\multirow{2}{*}{Trajectory in Fig.~\ref{fig:SimPatrol}} & 5.8 & 0.20 & 0.15    \\ 
\cline{2-4}
& 2.9 & 0.42 & 0.12  \\
\hline
\end{tabular}
\label{table:CompLocErr}
\end{table}

\begin{figure}
  \centering
  \includegraphics[width=0.9\linewidth]{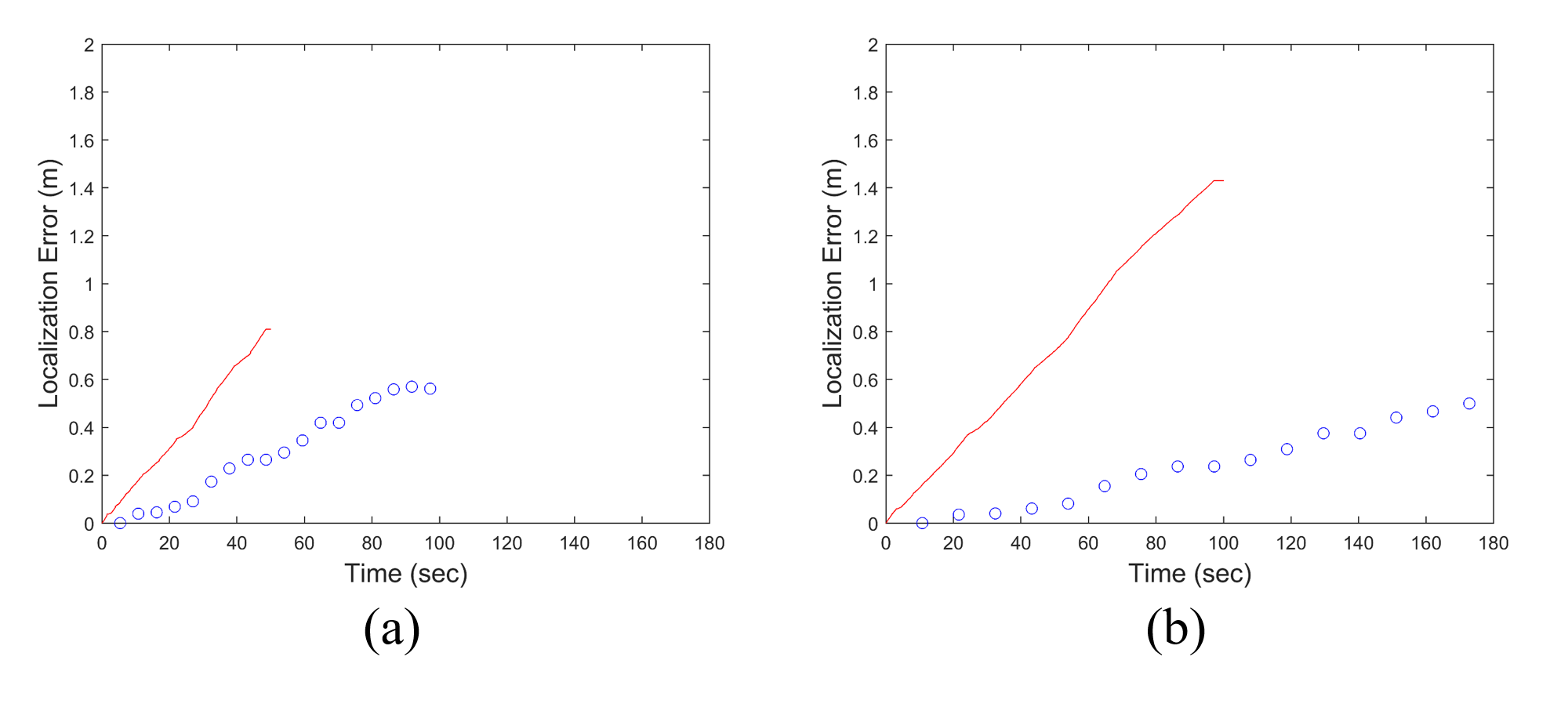}
  \caption{Comparison of localization errors with the dead-reckoning system (red curve) and the proposed system (blue circles) under the trajectory given in Fig.~\ref{fig:SimTrajTurns}.
  The intended wheel rotational speed, i.e., $\Omega_\mathrm{wheel}$ in Eq.~(\ref{eqn:WheelSpeed}), was set to (a) 5.8~rad/s and (b) 2.9~rad/s.}
\label{fig:ErrBehavior}
\end{figure}

It is interesting to note that the error accumulation behaviors of the two methods are markedly different.
Fig.~\ref{fig:ErrBehavior} compares these behaviors under the trajectory given in Fig.~\ref{fig:SimTrajTurns}. 
Fig.~\ref{fig:ErrBehavior}(a) shows the case where $\Omega_\mathrm{wheel} = 5.8$~rad/s, while Fig.~\ref{fig:ErrBehavior}(b) illustrates the case with $\Omega_\mathrm{wheel}$ reduced by half.

Because gyros measure the angular rate of rotation, the sensor measurements must be integrated over time to obtain the heading angle.
During this process, errors in the heading angle accumulate because sensor noise is also integrated.
As a result, the localization error of a dead-reckoning system using gyros increases over time.
When $\Omega_\mathrm{wheel}$ was reduced from 5.8~rad/s in Fig.~\ref{fig:ErrBehavior}(a) to 2.9~rad/s in Fig.~\ref{fig:ErrBehavior}(b), the travel time to the destination approximately doubled, which also doubled the duration of error accumulation in the dead-reckoning system.
Consequently, the red curve (i.e., the dead-reckoning system) in Fig.~\ref{fig:ErrBehavior}(b) shows approximately twice the localization error compared to the red curve in Fig.~\ref{fig:ErrBehavior}(a).
In other words, slower movement increases travel time and thereby increases localization error in a dead-reckoning system when following the same trajectory.

On the other hand, the localization error of the proposed system depends on the number of generated triangles en route to the destination and is independent of travel time.
The localization error can be evaluated only when the moving robot in Fig.~\ref{fig:MoveToTargetVertex} reaches the target vertex and a new triangle is formed.
Therefore, the localization errors of the proposed system over time are plotted as discrete points rather than a continuous curve in Fig.~\ref{fig:ErrBehavior}.

Recall that the moving robot finds the target vertex using the algorithm in Fig.~\ref{fig:TriFormAlgorithm}, and this algorithm is independent of the robot's speed.
Once a new triangle is formulated, the localization errors along the longitudinal and lateral axes---shown in Fig.~\ref{fig:TrajExample} and following the empirical CDFs in Fig.~\ref{fig:CDF}---are evaluated.
The localization error in the longitudinal axis is caused by the finite pixel size, which limits the resolution of lateral distance measurements (i.e., d\textsubscript{m1} and d\textsubscript{m2} in Fig.~\ref{fig:TriFormAlgorithm}), while the error in the lateral direction is primarily due to the nonholonomic constraint of the robots.
The number of generated triangles to reach the destination remains unchanged for a given trajectory, even if the robot speed varies.
Therefore, the localization errors of the proposed system at the destination in Fig.~\ref{fig:ErrBehavior}(a) and Fig.~\ref{fig:ErrBehavior}(b) are nearly identical.

\begin{figure}
  \centering
  \includegraphics[width=0.6\linewidth]{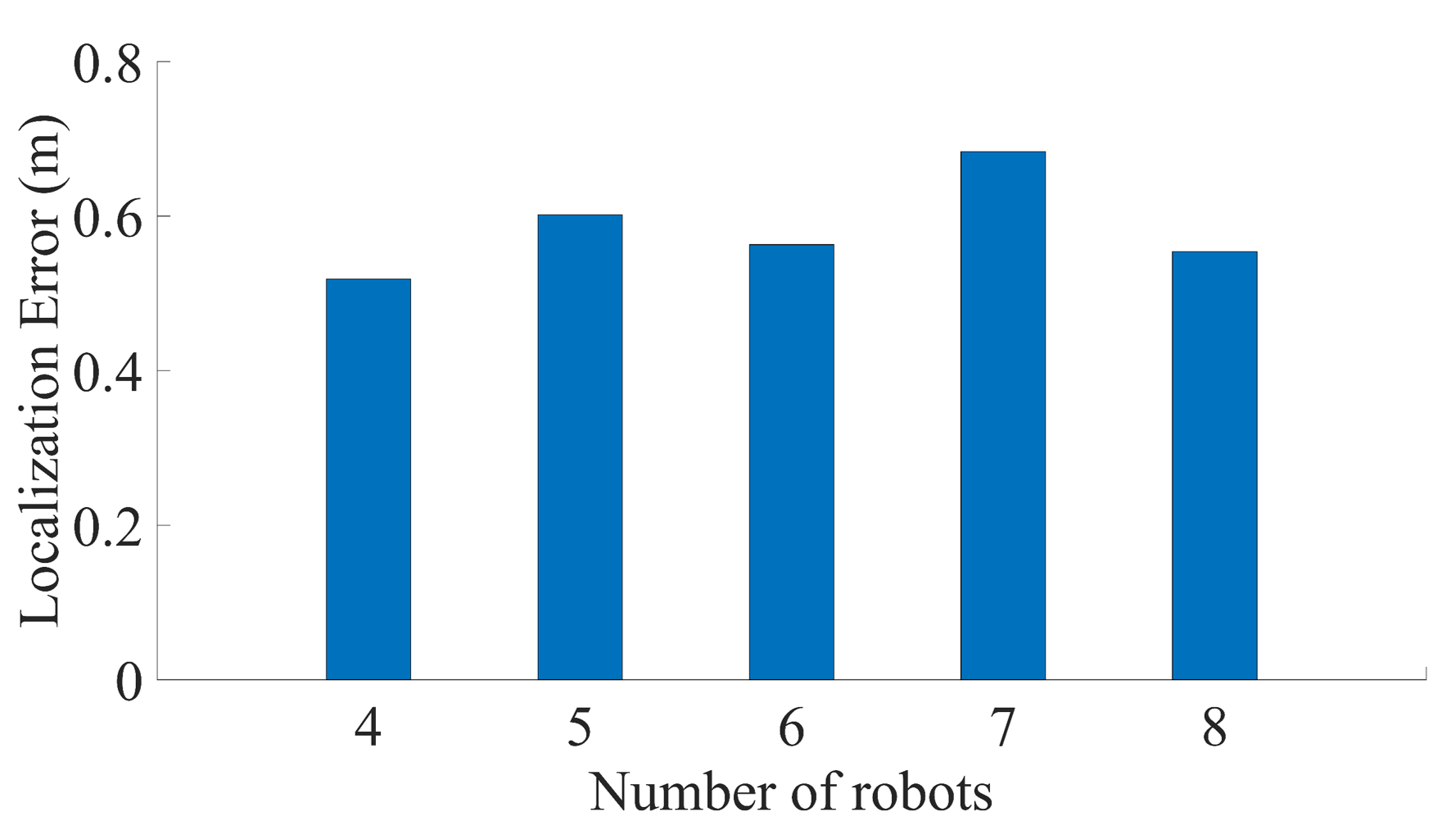}
  \caption{Localization error corresponding to the trajectory in Fig.~\ref{fig:SimTrajTurns}, evaluated for different numbers of robots to assess system scalability.}
\label{fig:scalAnalysis}
\end{figure}

Fig.~\ref{fig:scalAnalysis} shows the localization error when the trajectory used in Fig.~\ref{fig:SimTrajTurns} is executed with different numbers of robots for scalability analysis. 
(The reported errors represent average values based on 100 simulation runs.)
The initial positions of the robots and the selection of the moving robot follow the configuration shown in Fig.~\ref{fig:NrobotSystem}.
As the results indicate, the localization error remains consistent despite an increase in the number of robots, demonstrating good scalability even when extended to an $N$-robot system.
This is reasonable because the localization error in the proposed system depends on the number of generated triangles required to reach the destination.
If the distance to the destination is significantly greater than the size of the $N$-robot system itself---as is the case in Fig.~\ref{fig:SimTrajTurns}---the number of required triangles remains nearly constant, regardless of the number of robots.

\section{Conclusions}
\label{sec:conclusions}

This paper proposed a low-cost localization method suitable for multiple robots operating in wide open spaces.
In the proposed method, a minimum of four robots create and maintain equilateral triangular formations while progressing toward the destination.
Although the system relies only on low-cost monocular vision sensors and standard visual markers, a moving robot in the formation can accurately reach the target vertex with the assistance of beacon robots.
Since the depth information obtained from low-cost monocular vision sensors is typically not very accurate, the proposed system relies solely on lateral distance measurements and the geometric properties of equilateral triangles to achieve high localization accuracy.

We conducted experiments in a controlled laboratory environment using a motion capture system that provided ground truth trajectories of the robots.
Based on the experimental data, realistic error models of the proposed system were obtained.
The performance of the proposed system in wide open spaces---where SLAM is not particularly effective---was then compared with that of a dead-reckoning system through simulations.
The results confirm that the proposed system achieves noticeably higher accuracy than the dead-reckoning system, especially when travel time is extended.

Moreover, the error characteristics of the proposed system differ from those of the dead-reckoning system.
Unlike the dead-reckoning system, the localization error of the proposed system at the destination depends on the number of generated triangles along the path and is independent of travel time.
This property can be particularly advantageous for missions involving long travel durations.

\vspace{6pt} 

\authorcontributions{Conceptualization, T.K., J.-W.K., J.H.K.; methodology, T.K., J.H.K.; software, T.K., J.H.K.; validation, T.K., I.B., J.H.K.; resources, J.H.K.; data curation, T.K., I.B., J.H.K.; writing---original draft preparation, T.K.; writing---review and editing, J.H.K.; supervision, J.-W.K., J.H.K. All authors have read and agreed to the published version of the manuscript.}

\funding{This work was supported in part by Grant RS-2024-00407003 from the ``Development of Advanced Technology for Terrestrial Radionavigation System'' project, funded by the Ministry of Oceans and Fisheries, Republic of Korea; 
in part by the National Research Foundation of Korea (NRF), funded by the Korean government (Ministry of Science and ICT, MSIT), under Grant RS-2024-00358298;
and in part by the Institute of Information \& Communications Technology Planning \& Evaluation (IITP) through the Information Technology Research Center (ITRC) program, funded by the MSIT, under Grant IITP-2025-RS-2024-00437494.
}

\dataavailability{The data are contained within the article.} 

\acknowledgments{Generative AI (ChatGPT, OpenAI) was used solely to assist with grammar and language improvements during the manuscript preparation process.  
No content, ideas, data, or citations were generated by AI.  
All technical content, methodology, analysis, and conclusions were written and verified solely by the authors.}

\conflictsofinterest{The authors declare no conflicts of interest.}

\begin{adjustwidth}{-\extralength}{0cm}

\reftitle{References}


\bibliography{mybibfile, IUS_publications}

\begin{thebibliography}{999}

\bibitem[Pahlavan et~al.(2015)Pahlavan, Krishnamurthy, and Geng]{Pahlavan20153058}
Pahlavan, K.; Krishnamurthy, P.; Geng, Y.
\newblock Localization challenges for the emergence of the smart world.
\newblock {\em IEEE Access} {\bf 2015}, {\em 3},~3058--3067.
\newblock {\url{https://doi.org/10.1109/ACCESS.2015.2508648}}.

\bibitem[Fink et~al.(2013)Fink, Ribeiro, and Kumar]{Fink2013290}
Fink, J.; Ribeiro, A.; Kumar, V.
\newblock Robust control of mobility and communications in autonomous robot teams.
\newblock {\em IEEE Access} {\bf 2013}, {\em 1},~290--309.
\newblock {\url{https://doi.org/10.1109/ACCESS.2013.2262013}}.

\bibitem[Misra and Enge(2011)]{Enge11}
Misra, P.; Enge, P.
\newblock {\em Global Positioning System: Signals, Measurements, and Performance}; Ganga-Jamuna Press,  2011.

\bibitem[Tamazin et~al.(2016)Tamazin, Noureldin, Korenberg, and Massoud]{Tamazin201677}
Tamazin, M.; Noureldin, A.; Korenberg, M.; Massoud, A.
\newblock Robust fine acquisition algorithm for GPS receiver with limited resources.
\newblock {\em GPS Solut.} {\bf 2016}, {\em 20},~77--88.
\newblock {\url{https://doi.org/10.1007/s10291-015-0463-3}}.

\bibitem[Park and Seo(2021)]{Park21:Single}
Park, K.; Seo, J.
\newblock Single-antenna-based {GPS} antijamming method exploiting polarization diversity.
\newblock {\em IEEE Trans. Aerosp. Electron. Syst.} {\bf 2021}, {\em 57},~919--934.
\newblock {\url{https://doi.org/10.1109/TAES.2020.3034025}}.

\bibitem[Jia et~al.(2021)Jia, Lee, Khalife, Kassas, and Seo]{Jia21:Ground}
Jia, M.; Lee, H.; Khalife, J.; Kassas, Z.M.; Seo, J.
\newblock Ground vehicle navigation integrity monitoring for multi-constellation {GNSS} fused with cellular signals of opportunity.
\newblock In Proceedings of the IEEE ITSC,  2021, pp. 3978--3983.
\newblock {\url{https://doi.org/10.1109/ITSC48978.2021.9564686}}.

\bibitem[Lee et~al.(2022)Lee, Seo, and Kassas]{Lee22:Urban}
Lee, H.; Seo, J.; Kassas, Z.
\newblock Urban road safety prediction: A satellite navigation perspective.
\newblock {\em IEEE Intell. Transp. Syst. Mag.} {\bf 2022}, {\em 14},~94--106.
\newblock {\url{https://doi.org/10.1109/MITS.2022.3181557}}.

\bibitem[Kim et~al.(2023)Kim, Park, and Seo]{Kim23:Single}
Kim, S.; Park, S.; Seo, J.
\newblock Single antenna based {GPS} signal reception condition classification using machine learning approaches.
\newblock {\em J. Position. Navig. Timing} {\bf 2023}, {\em 12},~149--155.
\newblock {\url{https://doi.org/10.11003/JPNT.2023.12.2.149}}.

\bibitem[Lee et~al.(2023)Lee, Hwang, Ahn, Seo, and Park]{Lee23:Seamless}
Lee, Y.; Hwang, Y.; Ahn, J.Y.; Seo, J.; Park, B.
\newblock Seamless accurate positioning in deep urban area based on mode switching between {DGNSS} and multipath mitigation positioning.
\newblock {\em IEEE Trans. Intell. Transp. Syst.} {\bf 2023}, {\em 24},~5856--5870.
\newblock {\url{https://doi.org/10.1109/TITS.2023.3256040}}.

\bibitem[Kim and Seo(2023)]{Kim23:Machine}
Kim, S.; Seo, J.
\newblock Machine-learning-based classification of {GPS} signal reception conditions using a dual-polarized antenna in urban areas.
\newblock In Proceedings of the IEEE/ION PLANS,  Apr. 2023, pp. 113--118.
\newblock {\url{https://doi.org/10.1109/PLANS53410.2023.10140036}}.

\bibitem[Liu et~al.(2007)Liu, Darabi, Banerjee, and Liu]{Liu20071067}
Liu, H.; Darabi, H.; Banerjee, P.; Liu, J.
\newblock Survey of wireless indoor positioning techniques and systems.
\newblock {\em IEEE Trans. Syst. Man Cybern. Pt. C. Appl. Rev.} {\bf 2007}, {\em 37},~1067--1080.
\newblock {\url{https://doi.org/10.1109/TSMCC.2007.905750}}.

\bibitem[Lee et~al.(2020)Lee, Abdallah, Park, Seo, and Kassas]{Lee20:Neural}
Lee, H.; Abdallah, A.; Park, J.; Seo, J.; Kassas, Z.
\newblock Neural network-based ranging with {LTE} channel impulse response for localization in indoor environments.
\newblock In Proceedings of the ICCAS,  2020, pp. 939--944.
\newblock {\url{https://doi.org/10.23919/ICCAS50221.2020.9268386}}.

\bibitem[Lee et~al.(2022)Lee, Kang, Jeong, and Seo]{Lee22:Evaluation}
Lee, H.; Kang, T.; Jeong, S.; Seo, J.
\newblock Evaluation of {RF} fingerprinting-aided {RSS}-based target localization for emergency response.
\newblock In Proceedings of the IEEE VTC,  Jun. 2022, pp. 1--7.
\newblock {\url{https://doi.org/10.1109/VTC2022-Spring54318.2022.9860436}}.

\bibitem[Chen et~al.(2015)Chen, Meng, Wang, Zhang, Tian, and Yang]{Chen201524595}
Chen, G.; Meng, X.; Wang, Y.; Zhang, Y.; Tian, P.; Yang, H.
\newblock Integrated WiFi/PDR/smartphone using an unscented Kalman filter algorithm for 3D indoor localization.
\newblock {\em Sensors} {\bf 2015}, {\em 15},~24595--24614.
\newblock {\url{https://doi.org/10.3390/s150924595}}.

\bibitem[Guerra et~al.(2016)Guerra, Munguia, Bolea, and Grau]{Guerra2016}
Guerra, E.; Munguia, R.; Bolea, Y.; Grau, A.
\newblock Human collaborative localization and mapping in indoor environments with non-continuous stereo.
\newblock {\em Sensors} {\bf 2016}, {\em 16},~1--23.
\newblock {\url{https://doi.org/10.3390/s16030275}}.

\bibitem[Kang and Seo(2020)]{Kang20:Practical}
Kang, T.; Seo, J.
\newblock Practical simplified indoor multiwall path-loss model.
\newblock In Proceedings of the ICCAS,  2020, pp. 774--777.
\newblock {\url{https://doi.org/10.23919/ICCAS50221.2020.9268260}}.

\bibitem[Kim and Seo(2023)]{Kim23:Low}
Kim, W.; Seo, J.
\newblock Low-cost {GNSS} simulators with wireless clock synchronization for indoor positioning.
\newblock {\em IEEE Access} {\bf 2023}, {\em 11},~55861--55874.
\newblock {\url{https://doi.org/10.1109/ACCESS.2022.3210932}}.

\bibitem[Moon et~al.(2024)Moon, Park, and Seo]{Moon24:HELPS}
Moon, H.; Park, H.; Seo, J.
\newblock {HELPS} for emergency location service: Hyper-enhanced local positioning system.
\newblock {\em IEEE Wirel. Commun.} {\bf 2024}, {\em 31},~276--282.
\newblock {\url{https://doi.org/10.1109/MWC.011.2300354}}.

\bibitem[Lee et~al.(2022)Lee, Pullen, Lee, Park, Yoon, and Seo]{Lee22:Optimal}
Lee, H.; Pullen, S.; Lee, J.; Park, B.; Yoon, M.; Seo, J.
\newblock Optimal parameter inflation to enhance the availability of single-frequency {GBAS} for intelligent air transportation.
\newblock {\em IEEE Trans. Intell. Transp. Syst.} {\bf 2022}, {\em 23},~17801--17808.
\newblock {\url{https://doi.org/10.1109/TITS.2022.3157138}}.

\bibitem[Lee et~al.(2017)Lee, Morton, Lee, Moon, and Seo]{Lee17:Monitoring}
Lee, J.; Morton, Y.; Lee, J.; Moon, H.S.; Seo, J.
\newblock Monitoring and mitigation of ionospheric anomalies for {GNSS}-based safety critical systems.
\newblock {\em IEEE Signal Process. Mag.} {\bf 2017}, {\em 34},~96--110.
\newblock {\url{https://doi.org/10.1109/MSP.2017.2716406}}.

\bibitem[Sun et~al.(2020)Sun, Chang, Lee, Seo, Jade~Morton, and Pullen]{Sun20:Performance}
Sun, K.; Chang, H.; Lee, J.; Seo, J.; Jade~Morton, Y.; Pullen, S.
\newblock Performance benefit from dual-frequency {GNSS}-based aviation applications under Ionospheric Scintillation: {A} new approach to fading process modeling.
\newblock In Proceedings of the ION ITM,  2020, pp. 889--899.
\newblock {\url{https://doi.org/10.33012/2020.17184}}.

\bibitem[Sun et~al.(2021)Sun, Chang, Pullen, Kil, Seo, Morton, and Lee]{Sun21:Markov}
Sun, A.K.; Chang, H.; Pullen, S.; Kil, H.; Seo, J.; Morton, Y.J.; Lee, J.
\newblock Markov chain-based stochastic modeling of deep signal fading: Availability assessment of dual-frequency {GNSS}-based aviation under ionospheric scintillation.
\newblock {\em Space Weather} {\bf 2021}, {\em 19},~1--19.
\newblock {\url{https://doi.org/10.1029/2020SW002655}}.

\bibitem[Ahmed and Seo(2017)]{Ahmed17:Statistical}
Ahmed, N.; Seo, J.
\newblock Statistical evaluation of the multi-frequency {GPS} ionospheric scintillation observation data.
\newblock In Proceedings of the ICCAS,  2017, pp. 1792--1797.
\newblock {\url{https://doi.org/10.23919/ICCAS.2017.8204261}}.

\bibitem[Grant et~al.(2009)Grant, Williams, Ward, and Basker]{Grant2009173}
Grant, A.; Williams, P.; Ward, N.; Basker, S.
\newblock GPS jamming and the impact on maritime navigation.
\newblock {\em J. Navig.} {\bf 2009}, {\em 62},~173--187.
\newblock {\url{https://doi.org/10.1017/S0373463308005213}}.

\bibitem[Park et~al.(2018)Park, Lee, and Seo]{Park18:Dual}
Park, K.; Lee, D.; Seo, J.
\newblock Dual-polarized {GPS} antenna array algorithm to adaptively mitigate a large number of interference signals.
\newblock {\em Aerosp. Sci. Technol.} {\bf 2018}, {\em 78},~387--396.
\newblock {\url{https://doi.org/10.1016/j.ast.2018.04.029}}.

\bibitem[Chen et~al.(2012)Chen, Juang, Seo, Lo, Akos, {De Lorenzo}, and Enge]{Chen12:Design}
Chen, Y.H.; Juang, J.C.; Seo, J.; Lo, S.; Akos, D.; {De Lorenzo}, D.; Enge, P.
\newblock Design and implementation of real-time software radio for anti-interference {GPS/WAAS} sensors.
\newblock {\em Sensors} {\bf 2012}, {\em 12},~13417--13440.
\newblock {\url{https://doi.org/10.3390/s121013417}}.

\bibitem[Lee et~al.(2022)Lee, Kim, and Seo]{Lee22:SFOL}
Lee, S.; Kim, E.; Seo, J.
\newblock {SFOL DME} pulse shaping through digital predistortion for high-accuracy {DME}.
\newblock {\em IEEE Trans. Aerosp. Electron. Syst.} {\bf 2022}, {\em 58},~2616--2620.
\newblock {\url{https://doi.org/10.1109/TAES.2021.3123934}}.

\bibitem[Kim et~al.(2019)Kim, Park, and Seo]{Kim19:Mitigation}
Kim, S.; Park, K.; Seo, J.
\newblock Mitigation of {GPS} chirp jammer using a transversal {FIR} filter and {LMS} algorithm.
\newblock In Proceedings of the ITC-CSCC,  2019, pp. 1--4.
\newblock {\url{https://doi.org/10.1109/ITC-CSCC.2019.8793441}}.

\bibitem[Chen et~al.(2010)Chen, Juang, {De Lorenzo}, Seo, Lo, Enge, and Akos]{Chen10:Real}
Chen, Y.H.; Juang, J.C.; {De Lorenzo}, D.; Seo, J.; Lo, S.; Enge, P.; Akos, D.
\newblock Real-time software receiver for {GPS} controlled reception pattern antenna array processing.
\newblock In Proceedings of the ION GNSS,  2010, pp. 1932--1941.

\bibitem[Jubaer~Alam et~al.(2019)Jubaer~Alam, Ahamed, Faruque, Islam, and Tamim]{JubaerAlam2019}
Jubaer~Alam, M.; Ahamed, E.; Faruque, M.; Islam, M.; Tamim, A.
\newblock Left-handed metamaterial bandpass filter for GPS, Earth exploration-satellite and WiMAX frequency sensing applications.
\newblock {\em PLOS ONE} {\bf 2019}, {\em 14},~1--19.
\newblock {\url{https://doi.org/10.1371/journal.pone.0224478}}.

\bibitem[Kim et~al.(2022)Kim, Son, Park, Park, and Seo]{Kim22:First}
Kim, W.; Son, P.W.; Park, S.G.; Park, S.H.; Seo, J.
\newblock First demonstration of the {Korean eLoran} accuracy in a narrow waterway using improved {ASF} maps.
\newblock {\em IEEE Trans. Aerosp. Electron. Syst.} {\bf 2022}, {\em 58},~1492--1496.
\newblock {\url{https://doi.org/10.1109/TAES.2021.3114272}}.

\bibitem[Jirawimut et~al.(2003)Jirawimut, Ptasinski, Garaj, Cecelja, and Balachandran]{Jirawimut2003209}
Jirawimut, R.; Ptasinski, P.; Garaj, V.; Cecelja, F.; Balachandran, W.
\newblock A method for dead reckoning parameter correction in pedestrian navigation system.
\newblock {\em IEEE Trans. Instrum. Meas.} {\bf 2003}, {\em 52},~209--215.
\newblock {\url{https://doi.org/10.1109/TIM.2002.807986}}.

\bibitem[Cho et~al.(2011)Cho, Moon, Seo, and Baek]{Cho20112907}
Cho, B.S.; Moon, W.; Seo, W.J.; Baek, K.R.
\newblock A dead reckoning localization system for mobile robots using inertial sensors and wheel revolution encoding.
\newblock {\em J. Mech. Sci. Technol.} {\bf 2011}, {\em 25},~2907--2917.
\newblock {\url{https://doi.org/10.1007/s12206-011-0805-1}}.

\bibitem[Byun et~al.(2019)Byun, Lee, Han, Kim, Choi, and Kim]{Byun2019}
Byun, S.; Lee, H.; Han, J.; Kim, J.; Choi, E.; Kim, K.
\newblock Walking-speed estimation using a single inertial measurement unit for the older adults.
\newblock {\em PLOS ONE} {\bf 2019}, {\em 14},~1--16.
\newblock {\url{https://doi.org/10.1371/journal.pone.0227075}}.

\bibitem[Kim and Kwon(2023)]{Kim2023:Balancing}
Kim, Y.; Kwon, S.
\newblock Balancing-prioritized anti-slip control of a two-wheeled inverted pendulum robot vehicle on low-frictional surfaces with an acceleration slip indicator.
\newblock {\em Machines} {\bf 2023}, {\em 11},~1--19.
\newblock {\url{https://doi.org/10.3390/machines11050553}}.

\bibitem[Filliat and Meyer(2003)]{Filliat2003243}
Filliat, D.; Meyer, J.A.
\newblock Map-based navigation in mobile robots: I. A review of localization strategies.
\newblock {\em Cogn. Sys. Res.} {\bf 2003}, {\em 4},~243--282.
\newblock {\url{https://doi.org/10.1016/S1389-0417(03)00008-1}}.

\bibitem[Gamini~Dissanayake et~al.(2001)Gamini~Dissanayake, Newman, Clark, Durrant-Whyte, and Csorba]{GaminiDissanayake2001229}
Gamini~Dissanayake, M.; Newman, P.; Clark, S.; Durrant-Whyte, H.; Csorba, M.
\newblock A solution to the simultaneous localization and map building (SLAM) problem.
\newblock {\em IEEE Trans. Rob. Autom.} {\bf 2001}, {\em 17},~229--241.
\newblock {\url{https://doi.org/10.1109/70.938381}}.

\bibitem[Tang et~al.(2017)Tang, Zhang, Zou, and Liu]{Tang2017}
Tang, J.; Zhang, S.; Zou, Y.; Liu, F.
\newblock An adaptive map-matching algorithm based on hierarchical fuzzy system from vehicular GPS data.
\newblock {\em PLOS ONE} {\bf 2017}, {\em 12},~1--11.
\newblock {\url{https://doi.org/10.1371/journal.pone.0188796}}.

\bibitem[Thrun(2002)]{Thrun200252}
Thrun, S.
\newblock Probabilistic robotics.
\newblock {\em Commun. ACM} {\bf 2002}, {\em 45},~52--57.
\newblock {\url{https://doi.org/10.1145/504729.504754}}.

\bibitem[Durrant-Whyte and Bailey(2006)]{Durrant-Whyte200699}
Durrant-Whyte, H.; Bailey, T.
\newblock Simultaneous localization and mapping: Part I.
\newblock {\em IEEE Rob. Autom. Mag.} {\bf 2006}, {\em 13},~99--108.
\newblock {\url{https://doi.org/10.1109/MRA.2006.1638022}}.

\bibitem[Bresson et~al.(2015)Bresson, Feraud, Aufrere, Checchin, and Chapuis]{Bresson20151827}
Bresson, G.; Feraud, T.; Aufrere, R.; Checchin, P.; Chapuis, R.
\newblock Real-time monocular SLAM with low memory requirements.
\newblock {\em IEEE Trans. Intell. Transp. Syst.} {\bf 2015}, {\em 16},~1827--1839.
\newblock {\url{https://doi.org/10.1109/TITS.2014.2376780}}.

\bibitem[Munguia et~al.(2016)Munguia, Urzua, and Grau]{Munguia2016}
Munguia, R.; Urzua, S.; Grau, A.
\newblock Delayed monocular SLAM approach applied to unmanned aerial vehicles.
\newblock {\em PLOS ONE} {\bf 2016}, {\em 11},~1--24.
\newblock {\url{https://doi.org/10.1371/journal.pone.0167197}}.

\bibitem[Xu et~al.(2025)Xu, Chen, Yin, Ma, and Guo]{Xu2025:Virtual}
Xu, H.; Chen, C.; Yin, Q.; Ma, C.; Guo, F.
\newblock Virtual and real occlusion processing method of monocular visual assembly scene based on {ORB-SLAM3}.
\newblock {\em Machines} {\bf 2025}, {\em 13},~1--18.
\newblock {\url{https://doi.org/10.3390/machines13030212}}.

\bibitem[Forster et~al.(2017)Forster, Zhang, Gassner, Werlberger, and Scaramuzza]{Forster2017249}
Forster, C.; Zhang, Z.; Gassner, M.; Werlberger, M.; Scaramuzza, D.
\newblock SVO: Semidirect visual odometry for monocular and multicamera systems.
\newblock {\em IEEE Trans. Rob.} {\bf 2017}, {\em 33},~249--265.
\newblock {\url{https://doi.org/10.1109/TRO.2016.2623335}}.

\bibitem[Mu et~al.(2020)Mu, Yao, Zheng, Chen, Wang, and Qi]{Mu2020157628}
Mu, L.; Yao, P.; Zheng, Y.; Chen, K.; Wang, F.; Qi, N.
\newblock Research on SLAM algorithm of mobile robot based on the fusion of 2D LiDAR and depth camera.
\newblock {\em IEEE Access} {\bf 2020}, {\em 8},~157628--157642.
\newblock {\url{https://doi.org/10.1109/ACCESS.2020.3019659}}.

\bibitem[Daoud et~al.(2018)Daoud, Sabri, Loo, and Mansoor]{Daoud2018}
Daoud, H.; Sabri, A.; Loo, C.; Mansoor, A.
\newblock SLAMM: Visual monocular SLAM with continuous mapping using multiple maps.
\newblock {\em PLOS ONE} {\bf 2018}, {\em 13},~1--22.
\newblock {\url{https://doi.org/10.1371/journal.pone.0195878}}.

\bibitem[Aznar et~al.(2014)Aznar, Pujol, Pujol, Rizo, and Pujol]{Aznar2014}
Aznar, F.; Pujol, F.; Pujol, M.; Rizo, R.; Pujol, M.J.
\newblock Learning probabilistic features for robotic navigation using laser sensors.
\newblock {\em PLOS ONE} {\bf 2014}, {\em 9},~1--21.
\newblock {\url{https://doi.org/10.1371/journal.pone.0112507}}.

\bibitem[Chen et~al.(2023)Chen, Wang, Gao, Shang, Zhou, Li, Xu, and Hu]{Chen2023:Overview}
Chen, W.; Wang, X.; Gao, S.; Shang, G.; Zhou, C.; Li, Z.; Xu, C.; Hu, K.
\newblock Overview of multi-robot collaborative {SLAM} from the perspective of data fusion.
\newblock {\em Machines} {\bf 2023}, {\em 11},~1--74.
\newblock {\url{https://doi.org/10.3390/machines11060653}}.

\bibitem[Shao et~al.(2023)Shao, Zhao, Chen, Yang, Chen, Feng, Zhang, and Teng]{Shao2023:Advancing}
Shao, H.; Zhao, Q.; Chen, H.; Yang, W.; Chen, B.; Feng, Z.; Zhang, J.; Teng, H.
\newblock Advancing simultaneous localization and mapping with multi-sensor fusion and point cloud de-distortion.
\newblock {\em Machines} {\bf 2023}, {\em 11},~1--19.
\newblock {\url{https://doi.org/10.3390/machines11060588}}.

\bibitem[Williams et~al.(2009)Williams, Cummins, Neira, Newman, Reid, and Tardós]{Williams20091188}
Williams, B.; Cummins, M.; Neira, J.; Newman, P.; Reid, I.; Tardós, J.
\newblock A comparison of loop closing techniques in monocular SLAM.
\newblock {\em Rob. Autom. Syst.} {\bf 2009}, {\em 57},~1188--1197.
\newblock {\url{https://doi.org/10.1016/j.robot.2009.06.010}}.

\bibitem[Newman and Ho(2005)]{Newman2005635}
Newman, P.; Ho, K.
\newblock SLAM--Loop closing with visually salient features.
\newblock In Proceedings of the IEEE ICRA,  2005, pp. 635--642.
\newblock {\url{https://doi.org/10.1109/ROBOT.2005.1570189}}.

\bibitem[Wang et~al.(2021)Wang, Gao, Ding, Zhang, and Cai]{Wang2021}
Wang, G.; Gao, S.; Ding, H.; Zhang, H.; Cai, H.
\newblock LIO-CSI: LiDAR inertial odometry with loop closure combined with semantic information.
\newblock {\em PLOS ONE} {\bf 2021}, {\em 16},~1--18.
\newblock {\url{https://doi.org/10.1371/journal.pone.0261053}}.

\bibitem[Kurazume et~al.(1994)Kurazume, Nagata, and Hirose]{Kurazume19941250}
Kurazume, R.; Nagata, S.; Hirose, S.
\newblock Cooperative positioning with multiple robots.
\newblock In Proceedings of the IEEE ICRA,  1994, pp. 1250--1256.
\newblock {\url{https://doi.org/10.1109/ROBOT.1994.351315}}.

\bibitem[Kim et~al.(2017)Kim, Kwon, and Seo]{Kim17:Simulation}
Kim, J.; Kwon, J.W.; Seo, J.
\newblock Simulation study on a method to localize four mobile robots based on triangular formation.
\newblock In Proceedings of the ION Pacific PNT,  2017, pp. 348--361.
\newblock {\url{https://doi.org/10.33012/2017.15034}}.

\bibitem[Anousaki and Kyriakopoulos(1999)]{Anousaki199942}
Anousaki, G.; Kyriakopoulos, K.
\newblock Simultaneous localization and map building for mobile robot navigation.
\newblock {\em IEEE Rob. Autom. Mag.} {\bf 1999}, {\em 6},~42--53.
\newblock {\url{https://doi.org/10.1109/100.793699}}.

\bibitem[Guivant and Nebot(2001)]{Guivant2001242}
Guivant, J.; Nebot, E.
\newblock Optimization of the simultaneous localization and map-building algorithm for real-time implementation.
\newblock {\em IEEE Trans. Rob. Autom.} {\bf 2001}, {\em 17},~242--257.
\newblock {\url{https://doi.org/10.1109/70.938382}}.

\bibitem[Mourikis and Roumeliotis(2006)]{Mourikis20061273}
Mourikis, A.; Roumeliotis, S.
\newblock Predicting the performance of cooperative simultaneous localization and mapping (C-SLAM).
\newblock {\em Int. J. Rob. Res.} {\bf 2006}, {\em 25},~1273--1286.
\newblock {\url{https://doi.org/10.1177/0278364906072515}}.

\bibitem[Poulose and Han(2019)]{Poulose2019}
Poulose, A.; Han, D.
\newblock Hybrid indoor localization using IMU sensors and smartphone camera.
\newblock {\em Sensors} {\bf 2019}, {\em 19},~1--17.
\newblock {\url{https://doi.org/10.3390/s19235084}}.

\bibitem[Rhee and Seo(2018)]{Rhee18:Ground}
Rhee, J.; Seo, J.
\newblock Ground reflection elimination algorithms for enhanced distance measurement to the curbs using ultrasonic sensors.
\newblock In Proceedings of the ION ITM,  2018, pp. 224--231.
\newblock {\url{https://doi.org/10.33012/2018.15552}}.

\bibitem[Rhee and Seo(2019)]{Rhee19:Low}
Rhee, J.; Seo, J.
\newblock Low-cost curb detection and localization system using multiple ultrasonic sensors.
\newblock {\em Sensors} {\bf 2019}, {\em 19},~1--22.
\newblock {\url{https://doi.org/10.3390/s19061389}}.

\bibitem[Kim et~al.(2016)Kim, Kwon, and Seo]{Kim16:Mapping}
Kim, J.; Kwon, J.W.; Seo, J.
\newblock Mapping and path planning using communication graph of unlocalized and randomly deployed robotic swarm.
\newblock In Proceedings of the ICCAS,  2016, Vol.~0, pp. 865--868.
\newblock {\url{https://doi.org/10.1109/ICCAS.2016.7832414}}.

\bibitem[Olson(2011)]{Olson2011:AprilTag}
Olson, E.
\newblock AprilTag: A robust and flexible visual fiducial system.
\newblock In Proceedings of the IEEE Int. Conf. Rob. Autom.,  2011, p. 3400 – 3407.
\newblock {\url{https://doi.org/10.1109/ICRA.2011.5979561}}.

\bibitem[Gebre-Egziabher(2004)]{Gebre-Egziabher14}
Gebre-Egziabher, D.
\newblock Design and performance analysis of a low-cost aided dead reckoning navigator.
\newblock {Ph.D.} dissertation, Stanford Univ., Stanford, CA, USA,  2004.

\bibitem[Vezinet(2014)]{Vezinet14}
Vezinet, J.
\newblock Study of future on-board GNSS/INS hybridization architectures.
\newblock {Ph.D.} dissertation, Univ. of Toulouse, Toulouse, France,  2014.

\bibitem[Kwon et~al.(2015)Kwon, Kim, and Seo]{Kwon15:Multiple}
Kwon, J.W.; Kim, J.; Seo, J.
\newblock Multiple leader candidate and competitive position allocation for robust formation against member robot faults.
\newblock {\em Sensors} {\bf 2015}, {\em 15},~10771--10790.
\newblock {\url{https://doi.org/10.3390/s150510771}}.

\end{thebibliography}

%


\PublishersNote{}
\end{adjustwidth}
\end{document}